\documentclass[10pt,twocolumn,letterpaper]{article}

\usepackage[pagenumbers]{cvpr}

\usepackage[dvipsnames]{xcolor}

\definecolor{cvprblue}{rgb}{0.21,0.49,0.74}

\usepackage{amssymb}
\usepackage{amsmath}
\usepackage{pifont}
\usepackage{bm}
\usepackage{booktabs}
\usepackage{mathtools}
\usepackage{multirow}

\usepackage[numbers]{natbib}
\usepackage[resetlabels]{multibib}
\newcites{supp}{References}
\usepackage[pagebackref,breaklinks,colorlinks,citecolor=cvprblue]{hyperref}
\usepackage{etoolbox}
\makeatletter
\patchcmd\Hy@backout{\@auxout}{\@mainaux}{}{\fail}
\patchcmd\Hy@backout{\@auxout}{\@mainaux}{}{\fail}
\makeatother

\newcommand{\myparagraph}[1]{\vspace{2mm}\noindent\textbf{#1}}
\usepackage[lined,ruled,linesnumbered]{algorithm2e}

\usepackage{balance}
\usepackage{xcolor}
\usepackage{color,colortbl}

\definecolor{col_nose}{RGB}{181,228,140}
\definecolor{col_mouth}{RGB}{255,183,3}
\definecolor{col_forehead}{RGB}{189,178,255}
\definecolor{col_cheek}{RGB}{144,224,239}
\definecolor{col_table}{RGB}{175,227,246} % green

\definecolor{mycolor1}{rgb}{0.85000,0.32500,0.09800}%
\definecolor{mycolor2}{rgb}{0.92900,0.69400,0.12500}%
\definecolor{mycolor3}{rgb}{0.49400,0.18400,0.55600}%
\definecolor{mycolor4}{rgb}{0.87843,0.76471,0.98824}%
\definecolor{mycolor5}{rgb}{0.46600,0.67400,0.18800}%
\definecolor{mycolor6}{rgb}{0.30100,0.74500,0.93300}%
\definecolor{mycolor7}{rgb}{0.00000,0.44700,0.74100}%

\definecolor{best_two}{RGB}{72,149,239} % table
\definecolor{best}{RGB}{179,11,0} % table

\title{3D Face Reconstruction with the Geometric Guidance of \\Facial Part Segmentation}

\author{Zidu Wang$^{1,2}$, Xiangyu Zhu$^{1,2}$\thanks{Corresponding author: Xiangyu Zhu}, Tianshuo Zhang$^{1,2}$, Baiqin Wang$^{1,2}$, Zhen Lei$^{1,2,3}$\\
    $^{1}$State Key Laboratory of Multimodal Artificial Intelligence Systems,\\ Institute of Automation, Chinese Academy of Sciences\\
    $^{2}$School of Artificial Intelligence, University of Chinese Academy of Sciences\\
    $^{3}$ Centre for Artificial Intelligence and Robotics, Hong Kong Institute of Science \& Innovation,\\ Chinese Academy of Sciences\\
    {\tt\small \{wangzidu2022, wangbaiqin2024\}@ia.ac.cn}, {\tt\small \{xiangyu.zhu, tianshuo.zhang, zlei\}@nlpr.ia.ac.cn}\\
}

\begin{document}

\maketitle

\begin{abstract}{
\hyphenpenalty=5000
\tolerance=100
3D Morphable Models (3DMMs) provide promising 3D face reconstructions in various applications. However, existing methods struggle to reconstruct faces with extreme expressions due to deficiencies in supervisory signals, such as sparse or inaccurate landmarks. Segmentation information contains effective geometric contexts for face reconstruction. Certain attempts intuitively depend on differentiable renderers to compare the rendered silhouettes of reconstruction with segmentation, which is prone to issues like local optima and gradient instability. In this paper, we fully utilize the facial part segmentation geometry by introducing Part Re-projection Distance Loss (PRDL). Specifically, PRDL transforms facial part segmentation into 2D points and re-projects the reconstruction onto the image plane. Subsequently, by introducing grid anchors and computing different statistical distances from these anchors to the point sets, PRDL establishes geometry descriptors to optimize the distribution of the point sets for face reconstruction. PRDL exhibits a clear gradient compared to the renderer-based methods and presents state-of-the-art reconstruction performance in extensive quantitative and qualitative experiments. Our project is available at \href{https://github.com/wang-zidu/3DDFA-V3}{https://github.com/wang-zidu/3DDFA-V3}. 
}
\end{abstract}

\begin{figure}[t]
\begin{center}
   \includegraphics[width=1\linewidth]{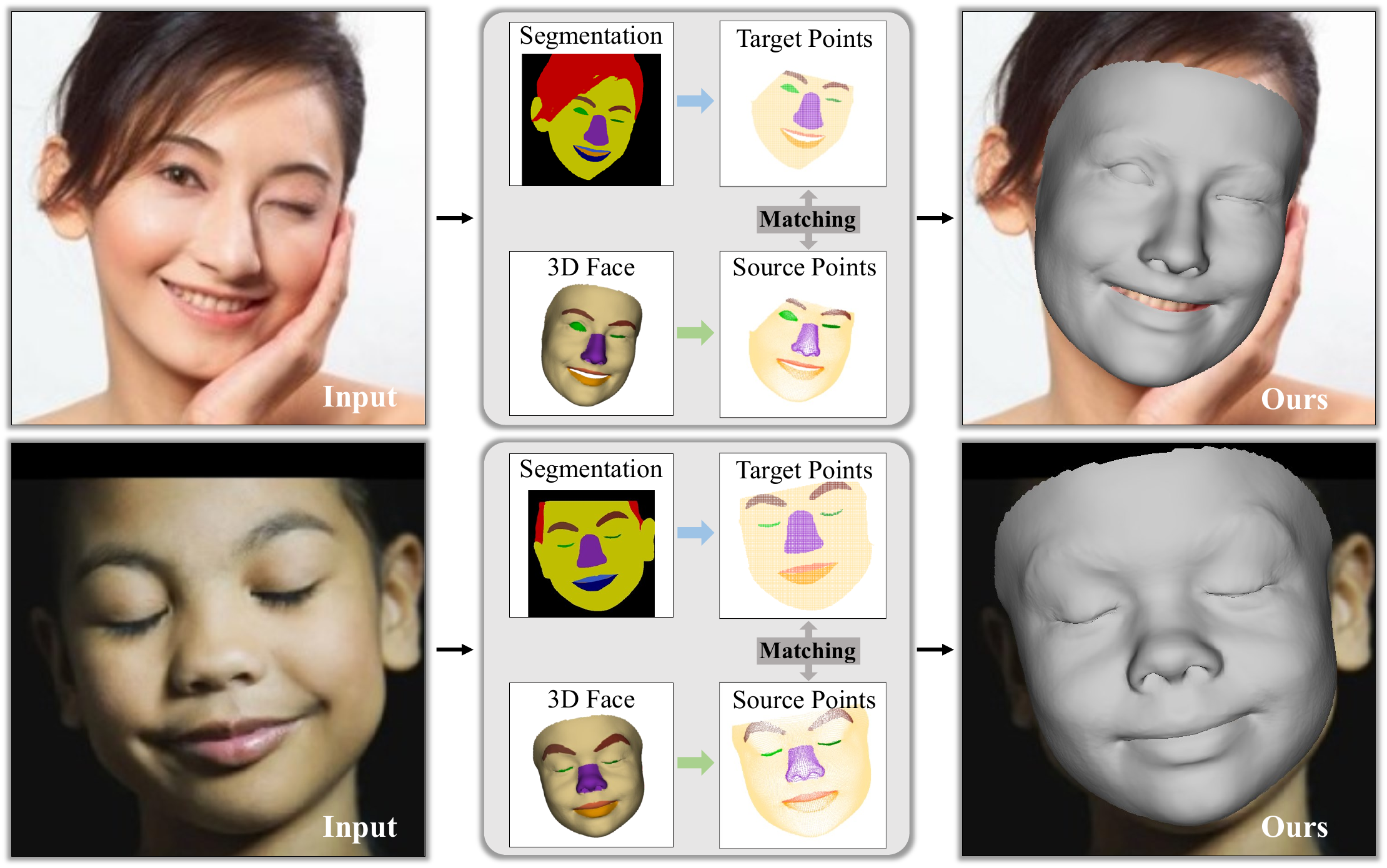}
\end{center}
\vspace{-0.35cm}
   \caption{We introduce Part Re-projection Distance Loss (PRDL) for 3D face reconstruction, leveraging the geometric guidance provided by facial part segmentation. PRDL enhances the alignment of reconstructed facial features with the original image and excels in capturing extreme expressions. 
   }
\label{shoutu}
 \vspace{-0.4cm}
\end{figure}

\section{Introduction}
Reconstructing 3D faces from 2D images is an essential task in computer vision and graphics, finding diverse applications in fields such as Virtual Reality (VR), Augmented Reality (AR), and Computer-generated Imagery (CGI), {\etc} In applications like VR makeup and AR emoji, 3DMMs \cite{blanz1999morphable} are commonly employed for precise facial feature positioning and capturing expressions. One of the most critical concerns is ensuring that the reconstructed facial components, including the eyes, eyebrows, lips, {\etc}, seamlessly align with their corresponding regions in the input image with pixel-level accuracy, particularly when dealing with extreme facial expressions, as shown in Fig.~\ref{shoutu}.

\begin{figure}[t]
\begin{center}
   \includegraphics[width=1\linewidth]{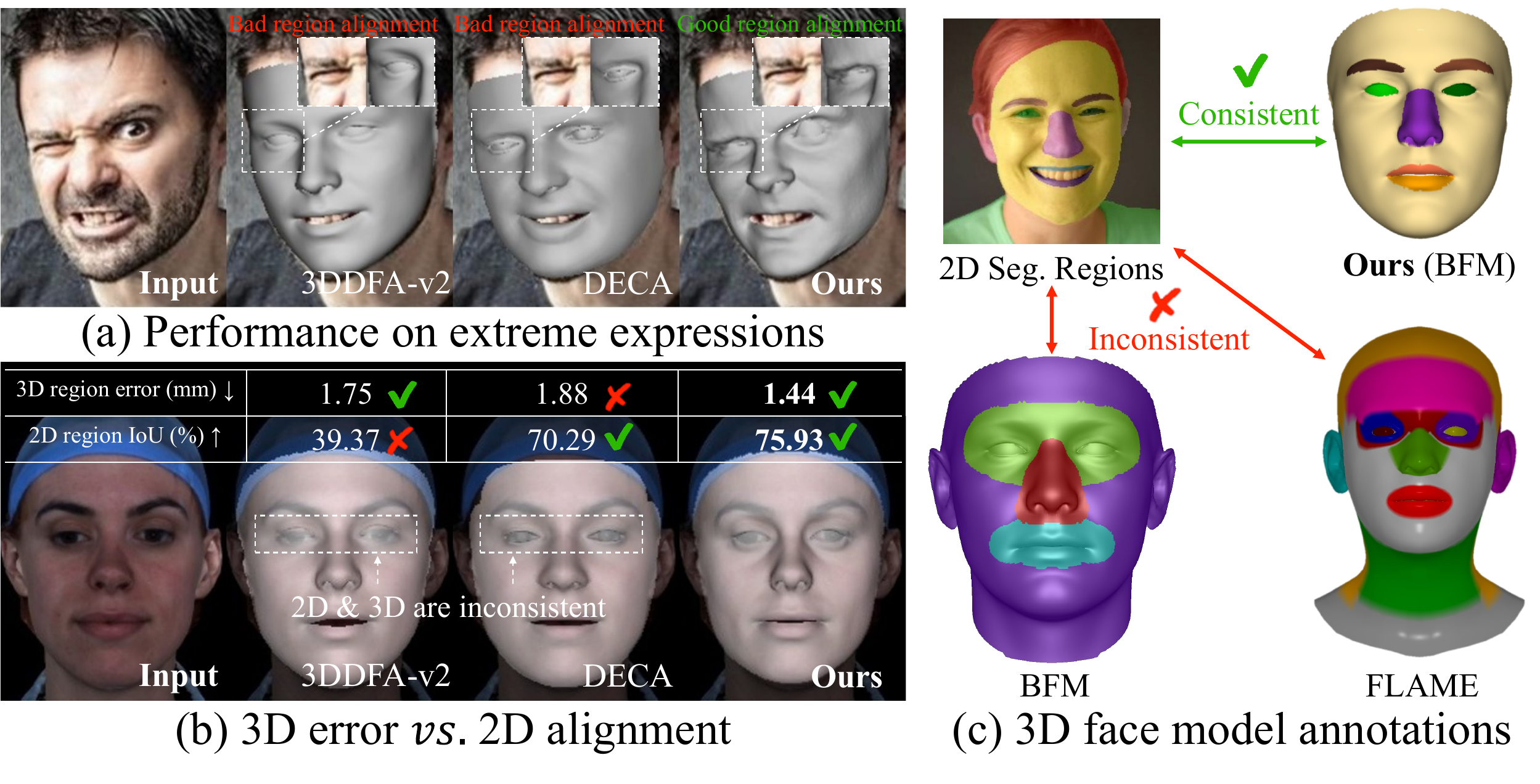}
\end{center}
\vspace{-0.7cm}
   \caption{Drawbacks of existing research and our results. (a) Present researches fail to reconstruct extreme expressions and perform bad region alignment. (b) Inconsistencies between 3D errors and 2D alignments, such as the eye region in this case. (c) Geometric optimization of each semantically consistent part is only achievable through our annotations.
   }
\label{second}
 \vspace{-0.4cm}
\end{figure}

Although current methods \cite{guo2020towards,deng2019accurate,DECA:Siggraph2021,lei2023hierarchical, xu10127617} have made notable strides in face reconstruction, some issues persist. On the one hand, existing works often rely on landmarks \cite{zielonka2022towards,guo2020towards} and photometric-texture \cite{shang2020self,egger2018occlusion} to guide face reconstruction. In the case of extreme facial expressions, landmarks are sparse or inaccurate and the gradient from the texture loss cannot directly constrain the shape \cite{zhu2022beyond}, posing a challenge for existing methods to achieve precise alignment of facial features in 3D face reconstruction, as depicted in Fig.~\ref{second}(a). On the other hand, many methods primarily adopt 3D errors as a quality metric, overlooking the precise alignment of facial parts. As shown in Fig.~\ref{second}(b), when evaluating the REALY \cite{chai2022realy} benchmark in the eye region, comparing the results of 3DDFA-v2 \cite{guo2020towards} and DECA \cite{DECA:Siggraph2021}, a lower 3D region error may not lead to better 2D region alignment. We believe in the potential for a more comprehensive utilization of the geometry information inherent in each facial part segmentation to guide 3D face reconstruction, addressing the issues mentioned above.

Facial part segmentation \cite{liu2020new,CelebAMask-HQ,lin2019face,lin2021roi} has been well developed, offering precise geometry for each facial feature with pixel-level accuracy. Compared with commonly used landmarks, part segmentation provides denser labels covering the whole image. Compared with photometric texture, part segmentation is less susceptible to lighting or shadow interference. Although facial part segmentation occasionally appears in the process of 3D face reconstruction, it is not fully utilized. For instance, it only serves to enhance the reconstruction quality of specific regions \cite{tewari2021learning, lei2023hierarchical}, or to distinguish the overall texture location for photometric-texture-loss \cite{li2021fit}, without delving into the specifics of facial parts. Attempts \cite{liu2019soft,zhu2020reda} to fit 3D parts with the guidance of segmentation information rely on differentiable renderers \cite{shreiner2009opengl, ravi2020pytorch3d, KaolinLibrary} to generate the silhouettes of the predicted 3D facial regions and optimize the difference between the rendered silhouettes and the 2D segmentation through Intersection over Union (IoU) loss. However, these renderers fail to provide sufficient and stable geometric signals for face reconstruction due to local optima, rendering error propagation, and gradient instability \cite{kato2020differentiable}.

This paper leverages the precise and rich geometric information in facial part silhouettes to guide face reconstruction, thereby improving the alignment of reconstructed facial features with the original image and excelling in reconstructing extreme facial expressions. Fig.\ref{shoutu} provides an overview of the proposed Part Re-projection Distance Loss (PRDL). Firstly, PRDL samples points within the segmented region and transforms the segmentation information into a 2D point set for each facial part. The 3D face reconstruction is also re-projected onto the image plane and transformed into 2D point sets for different regions. Secondly, PRDL samples the image grid anchors and establishes geometric descriptors. These descriptors are constructed by using various statistical distances from the anchors to the point set. Finally, PRDL optimizes the distribution of the same semantic point sets, leading to improved overlap between the regions covered by the target and predicted point sets. In contrast to renderer-based methods, PRDL exhibits a clear gradient. To facilitate the use of PRDL, we provide a new 3D mesh part annotation aligned with semantic regions in 2D face segmentation \cite{CelebAMask-HQ,Zheng2022DecoupledML}, which differs from the existing annotations \cite{bfm_seg_github,FLAME:SiggraphAsia2017}, as shown in Fig.\ref{second}(c). Besides the drawbacks of supervisory signals, the challenge of handling extreme expressions arises from data limitations. To boost studies and address the lack of emotional expression ({\eg}, closed-eye, open-mouth, frown, {\etc}), we synthesize a face dataset using the GAN-based method \cite{CelebAMask-HQ}. To highlight the performance of region overlapping, we propose a new benchmark to quantify the accuracy of 3D reconstruction parts cling to their corresponding image components on the 2D image plane. Our main contributions are as follows:

\begin{itemize}
\item We introduce a novel Part Re-projection Distance Loss (PRDL) to comprehensively utilize segmentation information for face reconstruction. PRDL transforms the target and prediction into semantic point sets, optimizing the distribution of point sets to ensure that the reconstructed regions and the target share the same geometry.

\item We introduce a new synthetic face dataset including closed-eye, open-mouth, and frown expressions, with more than $200K$ images.

\item Extensive experiments show that the results with PRDL achieve excellent performance and outperform the existing methods. The data and code are available at \href{https://github.com/wang-zidu/3DDFA-V3}{https://github.com/wang-zidu/3DDFA-V3}.

\end{itemize}

\section{Related Work}

\begin{figure*}[t]
\begin{center}
\includegraphics[width=1\linewidth]{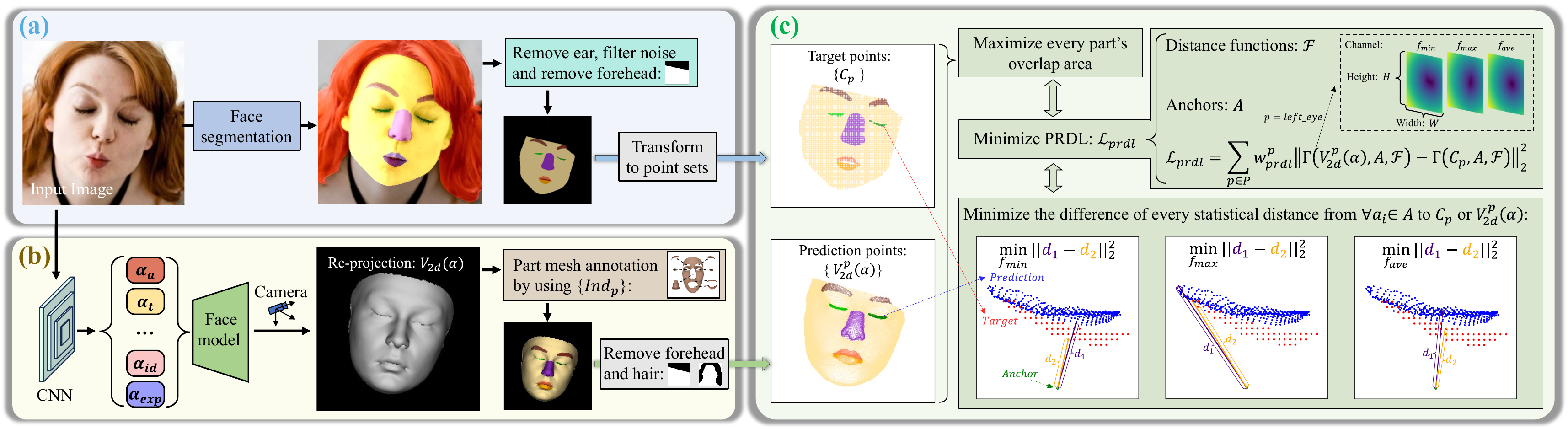 }
\end{center}

\vspace{-0.5cm}
\caption{Overview of Part Re-projection Distance Loss (PRDL). (a): Transforming facial part segmentation into target point sets $\{{\bm{C}_p}\}$. (b): Re-projecting ${V_{3d}}(\bm{\alpha} )$ onto the image plane to obtain predicted point sets $\{ {V_{2d}^p(\bm{\alpha} )\}} $. (c): Given anchors $\bm{A}$ and distance functions $\bm{\mathcal{F}}$, the core idea of PRDL is to minimize the difference of every statistical distance from any $\bm{a}_i \in \bm{A}$ to the ${V_{2d}^p(\bm{\alpha} )}$ or $\bm{C}_p$, leading to enhanced overlap between the regions covered by the target and predicted point sets.
   }
\label{our-method}
\vspace{-0.4cm}
\end{figure*}

\myparagraph{2D-to-3D Losses for 3D Face Reconstruction.} Landmark loss \cite{deng2019accurate,zielonka2022towards,guo2020towards} stands out as the most widely employed and effective supervised way for face reconstruction. Some studies \cite{dad3dheads,kartynnik2019real} reveal that it can generate 3D faces under the guidance of sufficient hundreds or thousands landmarks. Photometric loss is another commonly used loss involving rendering the reconstructed mesh with texture into an image and comparing it to the original input. Some researchers focus on predicting the facial features that need to be fitted while excluding occlusions \cite{shang2020self,egger2018occlusion}. The photometric loss is susceptible to factors like texture basis, skin masks, and rendering modes. It emphasizes overall visualization and may not effectively constrain local details. Perception loss \cite{genova2018unsupervised, DECA:Siggraph2021, deng2019accurate} distinguishes itself from image-level methods by employing pre-trained deep face recognition networks \cite{deng2019arcface} to extract high-level features from the rendered reconstruction results. These features are then compared with the features from the input. Lip segmentation consistency loss \cite{tewari2021learning} employs mouth segmentation to help reconstruction.

\myparagraph{Differentiable Silhouette Renderers.} The development of differentiable renderers \cite{shreiner2009opengl,ravi2020pytorch3d,KaolinLibrary} has enriched the supervised methods for 3D face reconstruction. These pipelines make the rasterization process differentiable, allowing for the computation of gradients for every pixel in the rendered results. By combining IoU loss with segmentation information, the silhouettes produced by these renderers have been shown to optimize 3D shapes \cite{liu2019soft,chen2019learning,zhu2020reda}. These rasterization processes typically rely on either local \cite{kato2018neural,loper2014opendr} or global \cite{liu2019soft,chen2019learning} geometric distance-based weighted aggregation, generating silhouettes by computing a probability related to the distance from pixels to mesh faces. However, to obtain a suitable sharp silhouette, the weight contribution of each position to the rendered pixel will decrease sharply with the increase of distance, and the gradient generated by the shape difference at the large distance will be small or zero, which makes it difficult to retain accurate geometry guidance. These renderers also encounter issues such as rendering error propagation and gradient instability \cite{kato2020differentiable}.

\myparagraph{Synthetic Dataset.} Synthetic data \cite{zhu2017face,raman2023mesh,wood2021fake} is commonly used to train 3D face reconstruction models \cite{guo2020towards,deng2019accurate,lei2023hierarchical}. However, these synthetic faces either prioritize the diversification of background, illumination, and identities \cite{raman2023mesh,wood2021fake}, or concentrate on pose variation \cite{zhu2017face}, contributing to achieve good results in reconstructing natural facial expressions but struggling to reconstruct extreme expressions. To overcome these limitations and facilitate the related research, this paper adopts a GAN-based method \cite{CelebAMask-HQ} to synthesize realistic and diverse facial expression data, including closed eyes, open mouths, and frowns.

\section{Methodology}
\subsection{Preliminaries}

We conduct a face model, an illumination model, and a camera model based on \cite{blanz2003face,guo2020towards,deng2019accurate,DECA:Siggraph2021}.

\myparagraph{Face Model.} The vertices and albedo of a 3D face is determined by the following formula:
\begin{equation}
\begin{small}
\begin{aligned}
{V_{3d}}(\bm{\alpha} ) &= \bm{R}({\bm{\alpha} _{a}})(\bm{\overline V}  + {\bm{\alpha} _{id}}{\bm{A}_{id}} + {\bm{\alpha} _{\exp }}{\bm{A}_{\exp }}) + {\bm{\alpha} _{t}}\\
{T_{alb}(\bm{\alpha} )} &= \bm{\overline T}  + {\bm{\alpha} _{alb}}{\bm{A}_{alb}}
\end{aligned},
\end{small}
\end{equation}
where ${{V_{3d}}(\bm{\alpha} ) \in {{\mathbb{R}}^{3 \times {35709}}}}$ is the 3D face vertices, $\bm{{\overline V }}$ is the mean shape. ${T_{alb}(\bm{\alpha} ) \in {{\mathbb{R}}^{3 \times {35709}}}}$ is the albedo, $\bm{{\overline T }}$ is the mean albedo. ${{\bm{A}_{id}}}$, ${{\bm{A}_{exp}}}$ and ${{\bm{A}_{alb}}}$ are the face identity vector bases, the expression vector bases and the albedo vector bases, respectively. ${{\bm{\alpha} _{id}} \in {{\mathbb{R}}^{{80}}}}$, ${{\bm{\alpha} _{exp}} \in {{\mathbb{R}}^{{64}}}}$ and ${{\bm{\alpha} _{alb}} \in {{\mathbb{R}}^{{80}}}}$ are the identity parameter, the expression parameter and the albedo parameter, respectively. ${{\bm{\alpha} _{t}} \in {{\mathbb{R}}^{{3}}}}$ is the translation parameter. ${\bm{R}({\bm{\alpha} _{a}}) \in {{\mathbb{R}}^{3 \times {3}}}}$ is the rotation matrix corresponding to \textit{pitch/raw/roll} angles ${{\bm{\alpha} _{a}} \in {{\mathbb{R}}^3}}$.

\myparagraph{Camera.} We employ a camera with a fixed perspective projection, which is same as \cite{deng2019accurate,lei2023hierarchical}. Using this camera to re-project ${{V_{3d}}(\bm{\alpha} )}$ into the 2D image plane yields ${{V_{2d}}(\bm{\alpha} ) \in {{\mathbb{R}}^{2 \times {35709}}}}$.

\myparagraph{Illumination Model.} Following \cite{DECA:Siggraph2021}, we adopt Spherical Harmonics (SH) \cite{ramamoorthi2001efficient} for the estimation of the shaded texture ${T_{tex}(\bm{\alpha})}$:
\begin{equation}
\begin{aligned}
\begin{array}{l}
{T_{tex}}(\bm{\alpha} ) = {T_{alb}}(\bm{\alpha} ) \odot \sum\limits_{k = 1}^9 {\bm{\alpha}_{sh}^k} {{\bm{\Psi }}_k}({\bm{N}})
\end{array}
\end{aligned},
\end{equation}
where $\odot$ denotes the Hadamard product, $\bm{N}$ is the surface normal of ${{V_{3d}}(\bm{\alpha} )}$, $\boldsymbol{\Psi}: \mathbb{R}^{3}\rightarrow \mathbb{R}$ is the SH basis function and ${{\bm{\alpha} _{sh}} \in {{\mathbb{R}}^{{9}}}}$ is the corresponding SH parameter. In summary, ${\bm{\alpha}  = [{\bm{\alpha} _{id}},{\bm{\alpha} _{\exp }},{\bm{\alpha} _{a}},{\bm{\alpha}_{t}}, {\bm{\alpha} _{sh}}]}$ is the undetermined parameter.

\subsection{Point Transformation on the Image Plane}\label{3-1}
\myparagraph{Transforming Segmentation to 2D Points.} For an input RGB face image ${\bm{I} \in {{\mathbb{R}}^{H \times W \times 3}}}$, the prediction of a face segmentation method can be represented by a set of binary tensors ${\bm{M} = \{ {\bm{M}_{p}}|{{p}} \in \bm{P} \} }$, where ${\bm{P}}=$ \{left\_eye, right\_eye, left\_eyebrow, right\_eyebrow, up\_lip, down\_lip, nose, skin\} and ${{\bm{M}_{p}} \in {\{ 0,1\} ^{H \times W}} }$. Specifically, ${\bm{M}_p^{(x,y)}=1}$ only if the 2D pixel position ${(x,y)}$ of ${\bm{M}_{p}}$ belongs to a certain face part ${p}$, and otherwise ${\bm{M}_p^{(x,y)} = 0}$. ${\bm{M}}$ can be transformed into a set of point sets ${{{\bm{C} = }}\{ {\bm{C}_p}|{{p}} \in \bm{P} \}}$, where ${{\bm{C}_p} = \{ (x,y)|\  if \ \bm{M}_p^{(x,y)} = 1\} }$. In this step, we employ DML-CSR \cite{Zheng2022DecoupledML} for face segmentation, excluding the ear regions, filtering out noise from the segmentation, and dynamically removing the forehead region above the eyebrows based on their position. This procedure is illustrated in Fig.~\ref{our-method}(a). More implementation details are provided in the supplemental materials. 

\myparagraph{Facial Part Annotation on 3D Face Model.} Our objective is to leverage $\{{\bm{C}_p}\}$ for guiding 3D face reconstruction. Thus, we should ensure that the reconstructed mesh can be divided into regions consistent with the semantics of the 2D segmentation. Due to the topological consistency of the face model, every vertex on the mesh can be annotated for a specific region. However, existing annotations \cite{FLAME:SiggraphAsia2017,bfm_seg_github,li2020learning} do not conform to widely accepted 2D face segmentation definitions \cite{CelebAMask-HQ,lin2021roi}, as shown in Fig.\ref{second}(c). To address this misalignment, we introduce new part annotations on both BFM \cite{blanz1999morphable} and FaceVerse \cite{wang2022faceverse}. We partition the vertices based on their indices. $i \in Ind_p$ indicates that the $i$-th vertex (denoted as $\bm{v}$) on the mesh belongs to part $p$. ${\{Ind_p|{{p}} \in \bm{P} \}}$ can be obtained by:
\begin{equation}
\begin{aligned}
\begin{array}{l}
{I^{seg}} = Seg(Render({V_{3d}} , Tex))\\
i \in In{d_p}, \ \  if \ \ {I^{seg}}(\bm{v}) \in p 
\end{array}
\end{aligned},
\end{equation}
where $Render( \cdot )$ generates an image by applying texture on the mesh, and $Seg(\cdot )$ is responsible for segmenting the rendered result. We employ different shape $V_{3d}$ and varying textures $Tex$ to label every $\bm{v} \in {V_{3d}} $ with hand-crafted modification. The annotation ${\{Ind_p\}}$ is pre-completed offline in the training process. Consequently, we utilize $\{{Ind_p}\}$ to transform the re-projection ${V_{2d}}(\bm{\alpha })$ into semantic point sets $\{ {V_{2d}^p(\bm{\alpha})}|{{p}} \in \bm{P} \}$. Besides, the upper forehead region situated above the eyebrows is dynamically excluded to ensure consistency with target. Points obstructed by hair are removed based on $\{{\bm{C}_p}\}$, as shown in Fig.~\ref{our-method}(b). Please refer to supplemental materials for annotation details.

\subsection{Part Re-projection Distance Loss (PRDL)}
\label{3-3}

This section describes the design of PRDL, focusing on constructing geometric descriptors and establishing the relation between the prediction $\{ {V_{2d}^p(\bm{\alpha} ) }\}$ and the ground truth $\{ {\bm{C}_p}\}$ for a given $p \in \bm{P}$, which is proved instrumental for face reconstruction.

In a more generalized formulation, considering two point sets $\bm{C} = \{ {\bm{c}_1},{\bm{c}_2},...,{\bm{c}_{|\bm{C}|}}\} $ and ${\bm{C}^*} = \{ \bm{c}_1^*,\bm{c}_2^*,...,\bm{c}_{|{\bm{C}^*}|}^*\} $, we aim to establish geometry descriptions by quantifying shape alignment between them for reconstruction. ${\bm{C}}$ and ${\bm{C}^*}$ may not possess the same number of points, and their points lack correspondence. Instead of directly searching the correspondence between the two sets, we use a set of fixed points as anchors $\bm{A} = \{ {\bm{a}_1},{\bm{a}_2},...,{\bm{a}_{|\bm{A}|}}\} $ and a collection of statistical distance functions $\bm{\mathcal{F}}= \{ {f_1},{f_2},...,{f_{|\bm{\mathcal{F}|}}}\} $ to construct geometry description tensors ${\bm{\Gamma} ({\bm{C},\bm{A},\bm{\mathcal{F}}})}\in {\mathbb{R}^{|\bm{A}| \times |\bm{\mathcal{F}}|}}$ and ${\bm{\Gamma}({{\bm{C}^*},\bm{A},\bm{\mathcal{F}}})}\in {\mathbb{R}^{|\bm{A}| \times |\bm{\mathcal{F}}|}}$ for ${\bm{C}}$ and ${\bm{C}^*}$, respectively (denoted as $\bm{\Gamma}$ and $\bm{\Gamma}^*$ for brevity). The value ${\bm{\Gamma}(i,j)}$ and ${\bm{\Gamma}^*(i,j)}$ at the position $(i,j)$ are determined by:
\begin{equation}{
\left\{
\begin{aligned}
{\bm{\Gamma}}(i,j) &= {f_j}({\bm{C}},{\bm{a}_i})\\
{\bm{\Gamma}^*}(i,j) &= {f_j}({\bm{C}^*},{\bm{a}_i})
\end{aligned},
\right.
}
\end{equation}
where every function ${f_j}(\bm{B},\bm{b}) \in \bm{\mathcal{F}}$ describes the distance from a single point $\bm{b}$ to a set of points $\bm{B}$, and ${f_j}(\bm{B},\bm{b})$ can be any statistically meaningful distance.

When fitting 3DMM to the segmented silhouettes for part $p$, we set $\bm{C}={V_{2d}^p(\bm{\alpha} )}$ and ${\bm{C}^*}=\bm{C}_p$ with specified anchors $\bm{A}$ and a set of distance functions $\bm{\mathcal{F}}$. Then we calculate their corresponding geometry descriptor tensors ${\bm{\Gamma}_p=\bm{\Gamma}({V_{2d}^p(\bm{\alpha} )},\bm{A},\bm{\mathcal{F}})}$ and ${\bm{\Gamma}_p^*}=\bm{\Gamma}(\bm{C}_p,\bm{A},\bm{\mathcal{F}})$. Part Re-projection Distance Loss (PRDL) $\mathcal{L}_{prdl}$ is defined as:
\begin{equation}
\begin{aligned}
\begin{array}{l}
{\mathcal{L}_{prdl}}  = \sum\limits_{p \in P} {w_{prdl}^p||{\bm{\Gamma}_p} - \bm{\Gamma}_p^*|{|_2^2}}
\end{array}
\end{aligned},
\end{equation}
where ${w_{prdl}^p}$ is the weight of each part $p$. In this paper, we set $\bm{\mathcal{F}}$ as a collection of the nearest ($f_{min}$), furthest ($f_{max}$), and average ($f_{ave}$) distance, {\ie} $ \bm{\mathcal{F}} = \{f_{max},f_{min},f_{ave}\}$. We set $\bm{A}$ as a $H\times W$ mesh grid. Then for ${\forall}{\bm{a}_i} \in \bm{A}$, the optimization objective of $\mathcal{L}_{prdl}$ is to:
\begin{equation}
\begin{aligned}
\begin{array}{l}
\left\{ {\begin{array}{*{20}{c}}
\min ||{f_{min }}({\bm{C}_p},{\bm{a}_i}) - {f_{min }}(V_{2d}^p(\bm{\alpha}),{\bm{a}_i})||_2^2\\
\min ||{f_{max }}({\bm{C}_p},{\bm{a}_i}) - {f_{max }}(V_{2d}^p(\bm{\alpha}),{\bm{a}_i})||_2^2\\
\min ||{f_{ave}}({\bm{C}_p},{\bm{a}_i}) - {f_{ave}}(V_{2d}^p(\bm{\alpha}),{\bm{a}_i})||_2^2
\end{array}} \right.
\end{array}
\end{aligned}.
\end{equation}

This process is shown in Fig.~\ref{our-method}(c). When $p=$ left\_eye, PRDL minimizes the length difference between the indigo and orange lines (also as shown in Fig.~\ref{grad}(a) when $p=$ right\_eyebrow). The upper right corner of Fig.~\ref{our-method}(c) is a visualization of $\bm{\Gamma}_{left\_eye}$ with the last channel separately by reshaping it from ${\mathbb{R}^{|\bm{A}| \times |\bm{\mathcal{F}}|}} $ to $ \mathbb{R}^{H\times W \times |\bm{\mathcal{F}}|}$. It is worth note that, the points number in ${V_{2d}^{p}(\bm{\alpha} )}$, $\bm{C}_{p}$ and $\bm{A}$ can be reduced by using Farthest Point Sampling (FPS) \cite{moenning2003fast} to decrease computational costs.

\subsection{Overall Losses}

To reconstruct a 3D face from image ${\bm{I} }$, we build frameworks to minimize the total loss $\mathcal{L}$ as follows:
\begin{equation}
\begin{aligned}
     \mathcal{L} &= \lambda_{{prdl}}{\mathcal{L}_{prdl}} + \lambda_{{lmk}}\mathcal{L}_{{lmk}}+  \lambda_{{pho}}\mathcal{L}_{{pho}}  \\ &+\lambda_{{per}}\mathcal{L}_{{per}} + \lambda_{reg}\mathcal{L}_{reg},
    \label{Eq.photo}
\end{aligned}
\end{equation}
where ${\mathcal{L}_{lmk}}$ is the landmark loss, we use detectors to locate $240$ 2D landmarks for ${\mathcal{L}_{lmk}}$ and adopt the dynamic landmark marching \cite{zhu2015high} to handle the non-correspondence between 2D and 3D cheek contour landmarks arising from pose variations. The photometric loss ${\mathcal{L}_{pho}}$ and the perceptual loss ${\mathcal{L}_{per}}$ are based on \cite{DECA:Siggraph2021,deng2019accurate}. ${\mathcal{L}_{reg}}$ is the regularization loss for $\bm{\alpha} $. $\lambda_{{prdl}}=0.8e-3$, $\lambda_{{lmk}}=1.6e-3$, $\lambda_{{pho}}=1.9$, $\lambda_{{per}}=0.2$, and $\lambda_{{reg}}=3e-4$ are the balance weights. ${\mathcal{L}_{prdl}}$ and ${\mathcal{L}_{lmk}}$ are normalized by ${H \times W}$.

\begin{figure}[t]
\begin{center}
   \includegraphics[width=1\linewidth]{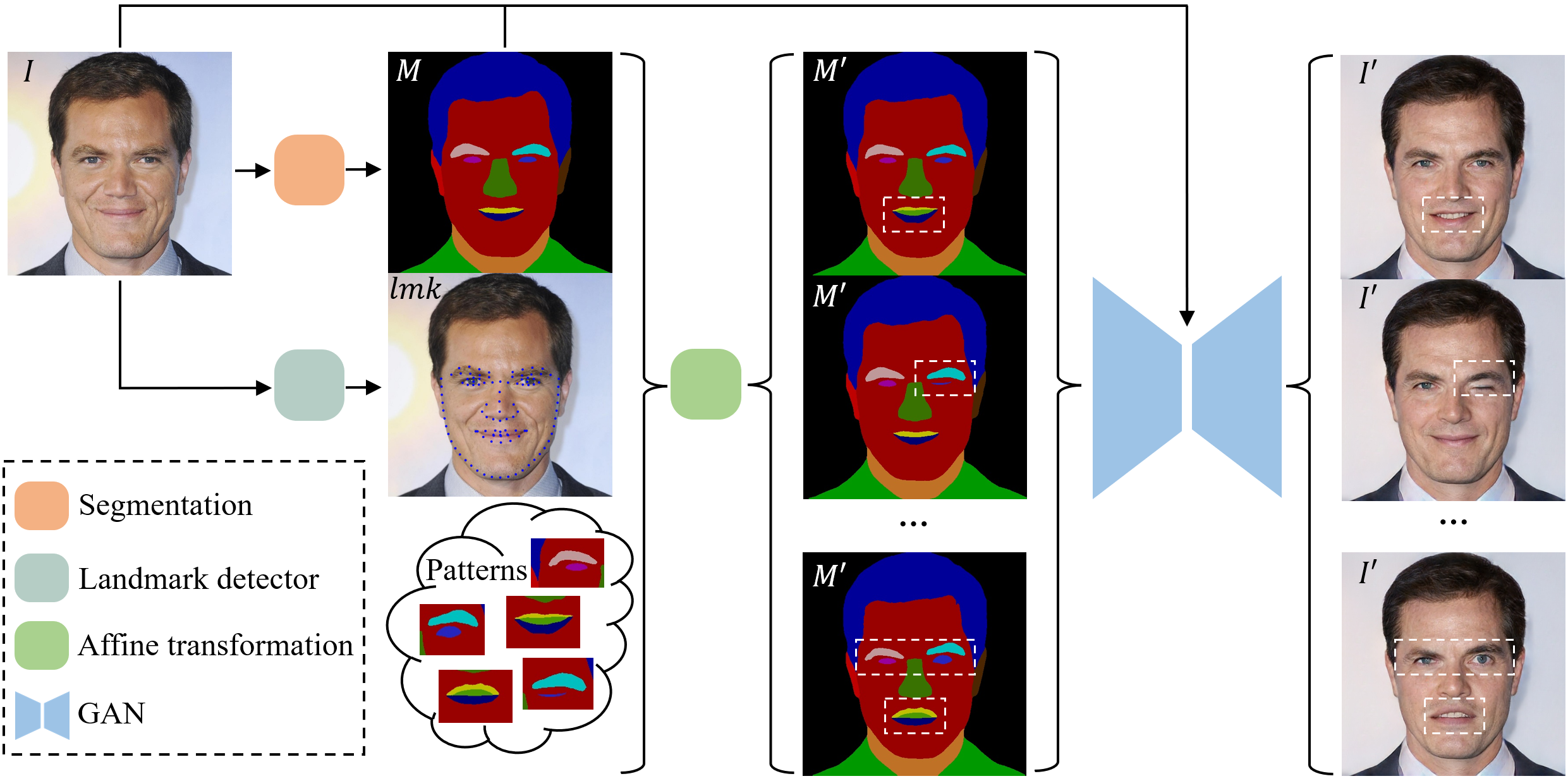}
\end{center}
\vspace{-0.6cm}
   \caption{Synthesize emotional expression data.
   }
\label{synthesis}
 \vspace{-0.2cm}
\end{figure}

\subsection{Synthetic Emotional Expression Data}
\begin{figure}[t]
\begin{center}
   \includegraphics[width=1\linewidth]{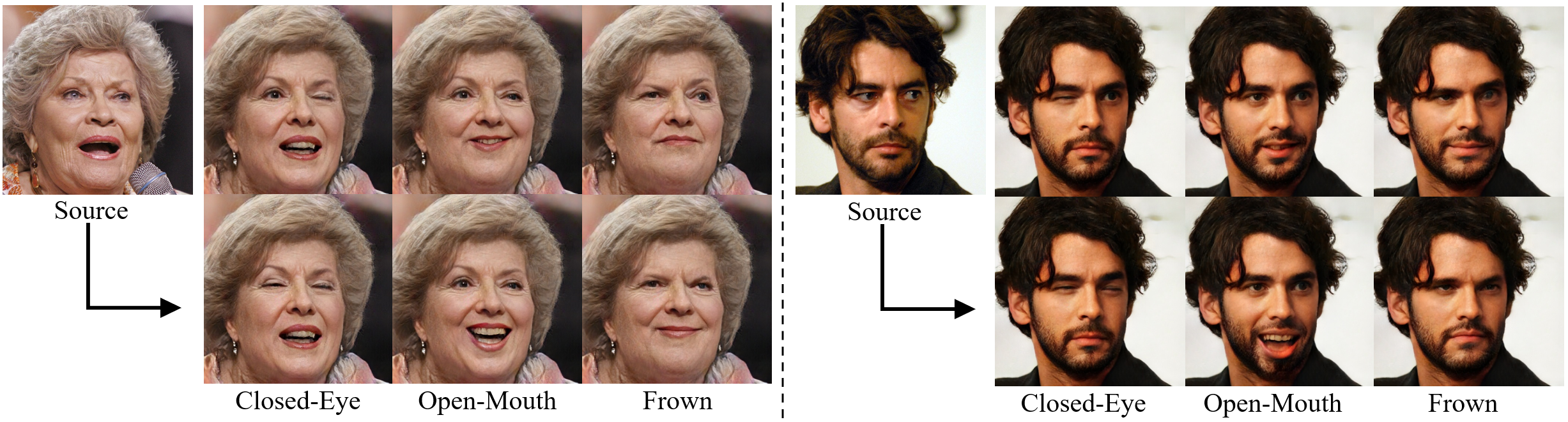}
\end{center} 
\vspace{-0.55cm}
   \caption{Examples of our synthetic face dataset.
   }
\label{synthesis-res}
 \vspace{-0.3cm}
\end{figure}

Benefiting from recent developments in face editing research \cite{CelebAMask-HQ,sun2022ide}, we can generate realistic faces through segmentation $\bm{M}$. We aim to mass-produce realistic and diverse facial expression data. To achieve this, we start by obtaining the segmentation $\bm{M}$ and landmarks ${lmk}$ of the original image $\bm{I}$ with a segmentation method \cite{Zheng2022DecoupledML} and a landmark detector, respectively. Leveraging the location of landmarks ${lmk}$, we apply affine transformation with various patterns onto the segmentation $\bm{M}$, resulting in ${\bm{M}'}$. Subsequently, ${\bm{M}'}$ is fed into the generative network \cite{CelebAMask-HQ} to produce a new facial expression image $\bm{I}'$, as depicted in Fig.~\ref{synthesis}. Based on CelebA \cite{liu2015faceattributes} and CelebAMask-HQ \cite{CelebAMask-HQ}, we have generated a dataset comprising more than $200K$ images, including expressions such as closed-eye, open-mouth, and frown, as depicted in Fig.~\ref{synthesis-res}. This dataset will be publicly available to facilitate research.

\begin{figure}[t]
\begin{center}
   \includegraphics[width=1\linewidth]{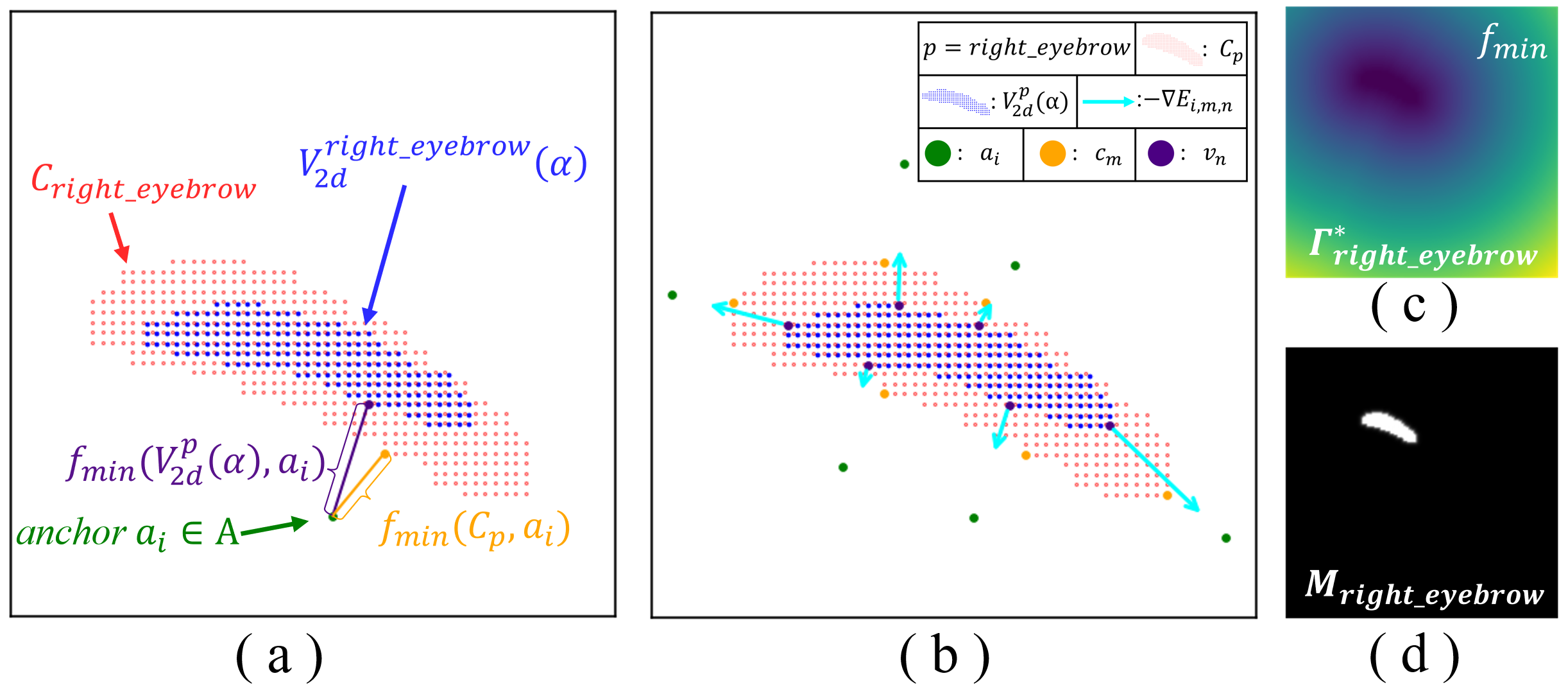}
\end{center}
 \vspace{-0.6cm}
   \caption{(a): $p=$ right\_eyebrow when the closest distance ($f_{min}$) is compared. (b): The gradient descent of PRDL for (a). (c): $\bm{\Gamma}_p^*$ is the regression target of PRDL in $f_{min}$ channel. (d): ${\bm{M}_{p}}$ is the regression target of renderer-based methods. $\bm{\Gamma}_p^*$ is more informative than ${\bm{M}_{p}}$.
   }
\label{grad}
 \vspace{-0.3cm}
\end{figure}

\begin{table*}
\scriptsize
\caption{Quantitative comparison on Part IoU benchmark. The best and runner-up are highlighted in \textbf{bold} and \underline{underlined}, respectively. R\_eye denotes the right eye, and similar definitions for the rest are omitted.
}
\vspace{-17pt}
\begin{center}
\resizebox{1\textwidth}{!}
{
\begin{tabular}{c|cccccccc}
\toprule[1pt]
\multirow{3}{*}{Methods} & \multicolumn{8}{c}{\textit{Part IoU(\%)$\uparrow$}}                                                                                                 \\ \cline{2-9} 
                         &  \textit{R\_eye}    & \textit{L\_eye}    & \textit{R\_brow}   & \textit{L\_brow}   & \textit{Nose}      & \textit{Up\_lip}   & \multicolumn{1}{c|}{\textit{Down\_lip}} &  \\
                         &  \textit{avg.$\pm$ std.}    & \textit{avg.$\pm$ std.}    & \textit{avg.$\pm$ std.}   & \textit{avg.$\pm$ std.}   & \textit{avg.$\pm$ std.}      & \textit{avg.$\pm$ std.}   & \multicolumn{1}{c|}{\textit{avg.$\pm$ std.}} &    \multirow{-2}{*}{\textit{avg.}}                  \\ 

\midrule[1pt]

PRNet \cite{feng2018joint}                      & 65.87$\pm$16.36  & 66.73$\pm$14.74  & 61.46$\pm$15.89   & 59.18$\pm$16.31   & 83.34$\pm$4.57 & 50.88$\pm$18.35   & \multicolumn{1}{c|}{58.16$\pm$17.72}     & 63.66 \\
MGCNet \cite{shang2020self}                     & 64.42$\pm$16.02  & 64.81$\pm$16.91  & 55.25$\pm$15.29   & 61.30$\pm$15.58   & 87.40\bm{$\pm$}3.51 & 41.16$\pm$19.70   & \multicolumn{1}{c|}{66.22$\pm$13.83}     & 62.94 \\
Deep3D \cite{deng2019accurate}                      & 71.87$\pm$12.00  & 70.52$\pm$12.19  & 64.66$\pm$11.31   & 64.70$\pm$11.98   & 87.69$\pm$3.51  & \underline{61.21$\pm$15.60}   & \multicolumn{1}{c|}{65.95$\pm$13.08}     & 69.51 \\
3DDFA-v2 \cite{guo2020towards}                 & 61.39$\pm$15.98  & 57.51$\pm$18.09  & 43.38$\pm$25.25   & 38.85$\pm$24.38   & 80.83$\pm$4.92 & 50.20$\pm$17.17   & \multicolumn{1}{c|}{59.01$\pm$15.23}     & 55.88 \\
HRN \cite{lei2023hierarchical}                  & 73.31$\pm$11.39   &  73.61$\pm$11.50   &  67.91$\pm$8.26   &  66.78$\pm$10.27   & \textbf{90.00}\bm{$\pm$}\textbf{2.60}  &  \textbf{63.80}\bm{$\pm$}\textbf{14.16}   & \multicolumn{1}{c|}{ 66.40$\pm$11.94  }     & 71.69  \\
DECA \cite{DECA:Siggraph2021}                  & 58.09$\pm$21.40  & 62.56$\pm$19.41  & 55.27$\pm$19.49   & 51.86$\pm$19.93   & 86.54$\pm$9.11 & 56.39$\pm$16.96   & \multicolumn{1}{c|}{62.81$\pm$17.66}     & 61.93 \\

Ours (w/o ${\mathcal{L}_{prdl}}$) & 70.72$\pm$9.44  & \underline{75.69$\pm$10.79}  & 71.11$\pm$8.58   & \underline{71.69$\pm$8.73}   & 88.35$\pm$4.60 & 57.26$\pm$15.97   & \multicolumn{1}{c|}{69.71$\pm$10.68}     & 72.08 \\
Ours (w/o Syn. Data) & \underline{73.81$\pm$10.12}  & 72.55$\pm$10.68  & \underline{72.24$\pm$9.23}   & 70.90$\pm$8.55   & 88.71$\pm$4.11  & 57.43$\pm$14.37    & \multicolumn{1}{c|}{\underline{69.87$\pm$10.54}}     & \underline{72.22} \\
\textbf{Ours}                     & \textbf{74.55}\bm{$\pm$}\textbf{11.46}  & \textbf{76.06}\bm{$\pm$}\textbf{10.32}  & \textbf{74.00}\bm{$\pm$}\textbf{7.72}   & \textbf{74.05}\bm{$\pm$}\textbf{7.70}   & \underline{89.06$\pm$3.53} & 58.16$\pm$12.76    & \multicolumn{1}{c|}{\textbf{70.86}\bm{$\pm$}\textbf{10.34 }}     & \textbf{73.82} \\ \bottomrule[1pt]
\end{tabular}
 }

\end{center}
\vspace{-10pt}
\vspace{-0.2cm}
\label{part-iou-tab}
\end{table*}

\begin{table*}
\scriptsize
\caption{Quantitative comparison on Realy benchmark. Lower values indicate better results. The best and runner-up are highlighted in \textbf{bold} and \underline{underlined}, respectively.
}
\vspace{-17pt}
\begin{center}
\renewcommand{\arraystretch}{1.05}
\resizebox{1\textwidth}{!}
{
\begin{tabular}{c|ccccc|ccccc}
\toprule[1pt]
                                                & \multicolumn{5}{c|}{  \textit{Frontal-view (mm) $\downarrow$}}                                                                                                  & \multicolumn{5}{c}{ \textit{Side-view (mm) $\downarrow$}}                                                                                                      \\  \cline{2-11}
                        & \textit{Nose}      & \textit{Mouth}     & \textit{Forehead}  & \multicolumn{1}{c|}{\textit{Cheek}}     &   & \textit{Nose}      & \textit{Mouth}     & \textit{Forehead}  & \multicolumn{1}{c|}{\textit{Cheek}}     &   \\
\multirow{-3}{*}{Methods}                       & \textit{avg.$\pm$ std.}     & \textit{avg.$\pm$ std.}     & \textit{avg.$\pm$ std.}  & \multicolumn{1}{c|}{\textit{avg.$\pm$ std.}}     & \multirow{-2}{*}{\textit{avg.}}   & \textit{avg.$\pm$ std.}      & \textit{avg.$\pm$ std.}     & \textit{avg.$\pm$ std.}  & \multicolumn{1}{c|}{\textit{avg.$\pm$ std.}}     & \multirow{-2}{*}{\textit{avg.}} \\

\midrule[1pt]

PRNet \cite{feng2018joint}     &   1.923$\pm$0.518        &   1.838$\pm$0.637        &   2.429$\pm$0.588        & \multicolumn{1}{c|}{1.863$\pm$0.698}          & \multicolumn{1}{c|}{2.013}     & 1.868$\pm$0.510          &   1.856$\pm$0.607        &  2.445$\pm$0.570         & \multicolumn{1}{c|}{1.960$\pm$0.731}          &  2.032    \\
MGCNet \cite{shang2020self}   &  1.771$\pm$0.380       & 1.417$\pm$0.409       & 2.268$\pm$0.503       & \multicolumn{1}{c|}{1.639$\pm$0.650} & \multicolumn{1}{c|}{1.774}                 & 1.827$\pm$0.383       & 1.409$\pm$0.418       & 2.248$\pm$0.508       & \multicolumn{1}{c|}{1.665$\pm$0.644} & 1.787                \\

Deep3D\cite{deng2019accurate} & 1.719$\pm$0.354       & 1.368$\pm$0.439       & 2.015$\pm$0.449       & \multicolumn{1}{c|}{1.528$\pm$0.501} & \multicolumn{1}{c|}{ 1.657}                 & 1.749$\pm$0.343       & 1.411$\pm$0.395      & 2.074$\pm$0.486       & \multicolumn{1}{c|}{1.528$\pm$0.517} & 1.691                \\

3DDFA-v2 \cite{guo2020towards} & 1.903$\pm$0.517       & 1.597$\pm$0.478       & 2.447$\pm$0.647       & \multicolumn{1}{c|}{1.757$\pm$0.642} & \multicolumn{1}{c|}{ 1.926 }                & 1.883$\pm$0.499       & 1.642$\pm$0.501       & 2.465$\pm$0.622       & \multicolumn{1}{c|}{1.781$\pm$0.636} & 1.943                \\
HRN \cite{lei2023hierarchical} &    1.722$\pm$0.330       &   {1.357$\pm$0.523}        &  1.995$\pm$0.476         & \multicolumn{1}{c|}{\textbf{1.072}\bm{$\pm$}\textbf{0.333}}          & \multicolumn{1}{c|}{1.537}     &  1.642$\pm$0.310         &   1.285$\pm$0.528        &   1.906$\pm$0.479        & \multicolumn{1}{c|}{\textbf{1.038}\bm{$\pm$}\textbf{0.322}}          & 1.468     \\
DECA \cite{DECA:Siggraph2021}  & 1.694$\pm$0.355       & 2.516$\pm$0.839       & 2.394$\pm$0.576       & \multicolumn{1}{c|}{1.479$\pm$0.535} & 2.010                 & 1.903$\pm$1.050       & 2.472$\pm$1.079       & 2.423$\pm$0.720       & \multicolumn{1}{c|}{1.630$\pm$1.135} & 2.107                \\

Ours (w/o ${\mathcal{L}_{prdl}}$) & 1.671$\pm$0.332       & 1.460$\pm$0.474       & 2.001$\pm$0.428       & \multicolumn{1}{c|}{1.142$\pm$0.315} & 1.568                 & 1.665$\pm$0.349       & 1.297$\pm$0.400       & 2.016$\pm$0.448       & \multicolumn{1}{c|}{1.134$\pm$0.342} & 1.528                \\

Ours (w/o Syn. Data) & \underline{1.592$\pm$0.327}       & \underline{1.339$\pm$0.433}       & \underline{1.823$\pm$0.407}       & \multicolumn{1}{c|}{1.119$\pm$0.332} & \underline{1.468}                 & \underline{1.628$\pm$0.320}       & \underline{1.229$\pm$0.433}       & \underline{1.872$\pm$0.407}       & \multicolumn{1}{c|}{1.091$\pm$0.312} & \underline{1.455}                \\

\textbf{Ours}                                              &    \textbf{1.586}\bm{$\pm$}\textbf{0.306}       &     \textbf{1.238}\bm{$\pm$}\textbf{0.373}      &     \textbf{1.810}\bm{$\pm$}\textbf{0.394}      & \multicolumn{1}{c|}{\underline{1.111$\pm$0.327}}          & \multicolumn{1}{c|}{\textbf{1.436}}     &    \textbf{1.623}\bm{$\pm$}\textbf{0.313}       &    \textbf{1.205}\bm{$\pm$}\textbf{0.366}       &       \textbf{1.864}\bm{$\pm$}\textbf{0.424}    & \multicolumn{1}{c|}{\underline{1.076$\pm$0.315}}          &   \textbf{1.442}   \\ 
\bottomrule[1pt]
\end{tabular}

 }
 
\end{center}
\vspace{-10pt}
 \vspace{-0.4cm}
 \label{realy-tab}
\end{table*}

\begin{figure*}[t]
\begin{center}
\includegraphics[width=1\linewidth]{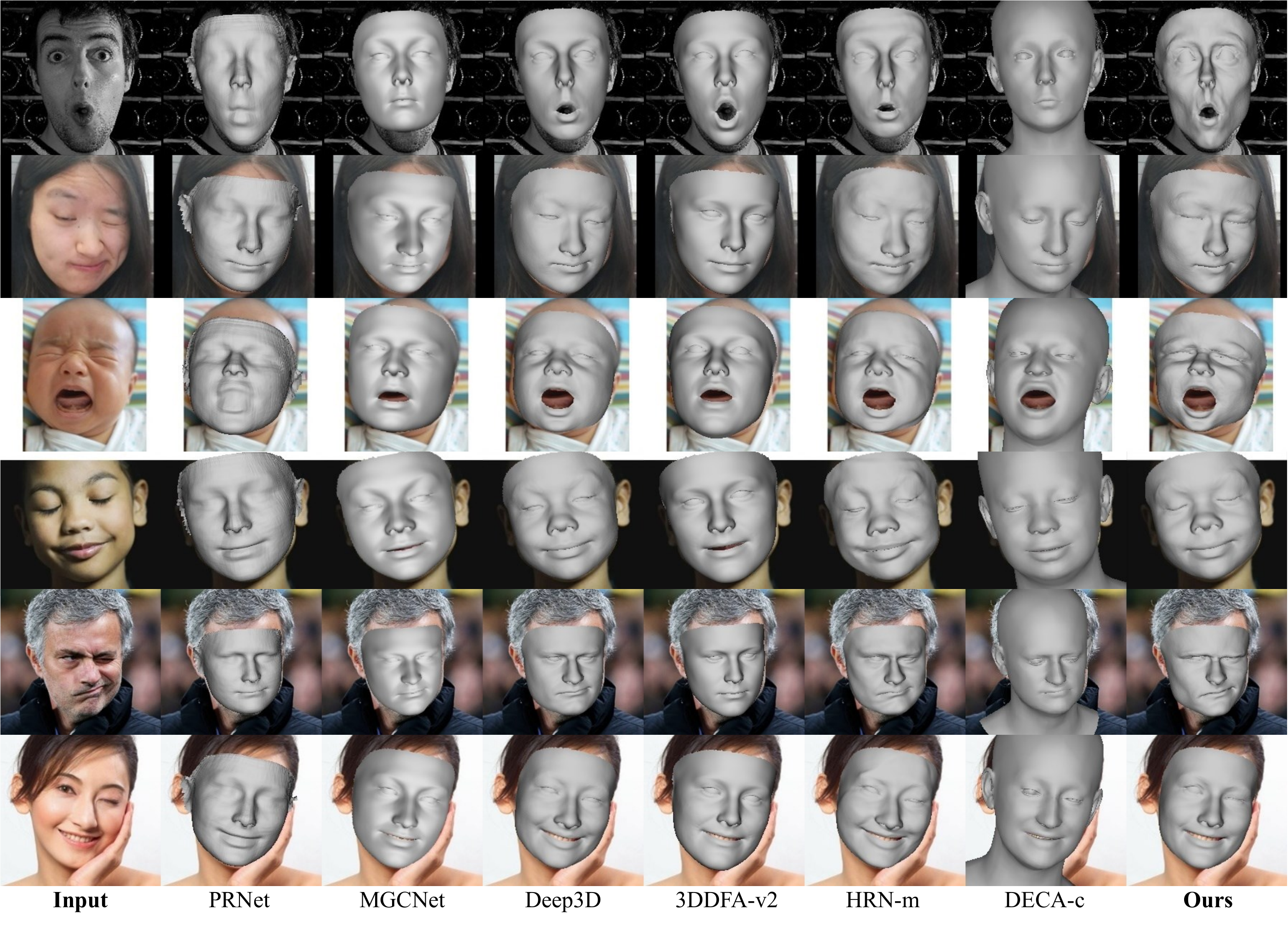 }
\end{center}
 \vspace{-0.75cm}
\caption{Qualitative comparison with the other methods. Our method achieves realistic reconstructions, particularly in the eye region.
}
    \vspace{-0.45cm}
\label{compare}
\end{figure*}

\section{Analysis of PRDL and Related Methods}

\myparagraph{The Gradient of PRDL.} With anchors and distance functions as the bridge, PRDL establishes the geometry descriptions of the two point sets. In Fig.~\ref{grad}, we take $p=$ right\_eyebrow as an example to analyze the gradient of PRDL. When considering $f_{min}$ and a specific anchor $\bm{a}_i \in \bm{A}$, $f_{min}$ identifies $\bm{c}_m$ and $\bm{v}_n$ from $\bm{C}_p$ and ${V_{2d}^p(\bm{\alpha} )}$, respectively, by selecting the ones closest to $\bm{a}_i$:

\begin{equation}
\label{argmin-c}
\begin{aligned}
\begin{array}{c}
m = \mathop {\arg \min }\limits_j ||{\bm{a}_i} - {\bm{c}_j}|{|_2},\;\;\;{\bm{c}_j} \in {\bm{C}_p},
\end{array}
\end{aligned}
\end{equation}
\begin{equation}
\label{argmin-v}
\begin{aligned}
\begin{array}{c}
n = \mathop {\arg \min }\limits_j ||{\bm{a}_i} - {\bm{v}_j}|{|_2},\;\;\;{\bm{v}_j} \in V_{2d}^p(\bm{\alpha} ).
\end{array}
\end{aligned}
\end{equation}

Under the definition of PRDL, the corresponding energy function $E_{i,m,n}$ for $\bm{a}_i$, $\bm{c}_m$ and $\bm{v}_n$ is:
\begin{equation}
\begin{aligned}
{E_{i,m,n}} &= {(||{\bm{a}_i} - {\bm{c}_m}|{|_2} - ||{\bm{a}_i} - {\bm{v}_n}|{|_2}  )}^2   \\
  \;\;\;\;\;\;\;\;\;\;\;  &= {({d_{i,m}} - {d_{i,n}})^2},
\end{aligned}
\end{equation}
where ${d_{i,m}} = ||{\bm{a}_i} - {\bm{c}_m}|{|_2}$, ${d_{i,n}} = ||{\bm{a}_i} - {\bm{v}_n}|{|_2}$. The gradient descent of ${E_{i,m,n}}$ on ${\bm{v}_n}$ is:
\begin{equation}
\label{equ-grad}
\begin{aligned}
\begin{array}{l}
{-\frac{{\partial {E_{i,m,n}}}}{{\partial {\bm{v}_n}}} = 2({\bm{v}_n} - {\bm{a}_i})(\frac{{{d_{i,m}}}}{{{d_{i,n}}}}-1)}.
\end{array}
\end{aligned}
\end{equation}

The physical explanation of Eqn.~\ref{equ-grad} is comprehensible and concise: the direction of $-\nabla {E_{i,m,n}}$ always aligns with the line connecting $\bm{a}_i$ and $\bm{v}_n$, if $d_{i,n}>d_{i,m}$, the direction of $-\nabla {E_{i,m,n}}$ is from $\bm{v}_n$ to $\bm{a}_i$ (as shown in Fig.~\ref{grad}(b)), and vice versa. In the context of gradient descent, the effect of $-\nabla {E_{i,m,n}}$ is to make $d_{i,n}=d_{i,m}$ as much as possible. Given $\bm{A}$ and $f_{min}$, the gradient descent of ${\mathcal{L}_{prdl}}$ on $\bm{v}_n$ is the aggregation of all anchors:
\begin{equation}
\label{equ-grad-all}
\begin{aligned}
\begin{array}{l}
-\frac{{\partial {\mathcal{L}_{prdl}}}}{{\partial {\bm{v}_n}}} =-w_{prdl}^p \sum\limits_{i,m} {\frac{{\partial {E_{i,m,n}}}}{{\partial {\bm{v}_n}}}}\\
 \;\;\;\;\; \;\;\;\;\;\;\;\;= -w_{prdl}^p  \sum\limits_{i,m} \nabla{E_{i,m,n}}.
\end{array}
\end{aligned}
\end{equation}

The scenario with $f_{max}$ is similar to that of $f_{min}$, with the only distinction lying in the selection of points. $f_{max}$ also has the capability to constrain ${V_{2d}^p(\bm{\alpha} )}$ within the confines of $\bm{C}_p$. $f_{ave}$ acts on the entire ${V_{2d}^p(\bm{\alpha} )}$, striving to bring its centroid as close as possible to the centroid of ${\bm{C}_{p}}$. The introduction of additional anchors and the integration of diverse statistical distances in PRDL prevent the optimization from local optima and provide sufficient geometric signals. Please refer to supplementary materials for more details.

\label{3-4}
\myparagraph{PRDL \vs Renderer-Based Loss: }An intuitive approach for fitting segmentation is to use the renderer-based IoU loss, where differentiable silhouette renderers play a crucial role. Consequently, we delve into the distinctions between PRDL and renderers. We can reshape $\bm{\Gamma}_p^*$ (${\mathbb{R}^{|\bm{A}| \times |\bm{\mathcal{F}}|}} \to  \mathbb{R}^{H\times W \times |\bm{\mathcal{F}}|}$) to visualize it with the last channel separately. Fig.~\ref{grad}(c) illustrates the visualization of the $f_{min}$ channel for $p=$ right\_eyebrow, while Fig.~\ref{grad}(d) represents the silhouette rendered by \cite{liu2019soft} or \cite{chen2019learning}. In comparison with the regression target $\bm{M}_p$ utilized in renderer-based methods, $\bm{\Gamma}_p^*$ in PRDL is more informative and more conducive to fitting. Please refer to supplementary materials for more details.

Furthermore, considering existing theoretical analyses \cite{chen2019learning,zhu2020reda,kato2020differentiable}, PRDL exhibits several notable advantages. First, in these renderers, all triangles constituting the object influence every pixel within the silhouettes, making it intricate to isolate specific geometric features. In contrast, $f_{min}$ or $f_{max}$ in PRDL matches the nearest or furthest point on the object, allowing for a more straightforward measurement of the shape's boundary characteristics. Secondly, these renderers either neglect pixels outside any triangles of the 3D object or assign minimal weights to them, emphasizing the rendered object region. However, this operation is equivalent to selectively choosing anchors $\bm{A}$ in the interior of the rendered shape, while the external anchors are either not chosen or treated differently by assigning small weights, thereby diminishing descriptive power. In Eqn.~\ref{equ-grad}, Eqn.~\ref{equ-grad-all} and Fig.~\ref{grad}(b), we have analyzed that external anchors play a significant role in the fitting process. Ablation study (Fig.\ref{compare-render}) also proves that PRDL is more effective than renderer-based methods like \cite{liu2019soft,chen2019learning,zhu2020reda}.

\section{Experiments}
\subsection{Experimental Settings}

\myparagraph{Reconstruction Frameworks. }We implement PRDL based on PyTorch \cite{paszke2019pytorch} and PyTorch3D~\cite{ravi2020pytorch3d}. We use ResNet-50 \cite{he2016deep} as the backbone to predict ${\bm{\alpha}}$. The input image is cropped and aligned by \cite{Deng2020CVPR}, and resized into $224\times224$.

\myparagraph{Data. }The face images are from publicly available datasets: Dad-3dheads \cite{dad3dheads}, CelebA \cite{liu2015faceattributes}, RAF-ML \cite{DBLP:journals/ijcv/ShangD19}, RAF-DB \cite{li2019reliable} and 300W \cite{sagonas2013300}. Our synthetic images are mainly from \cite{liu2015faceattributes, CelebAMask-HQ}. We use \cite{zhu2017face} for face pose augmentation. In total, our training data contained about $600K$ face images. We employ DML-CSR \cite{Zheng2022DecoupledML} to predict 2D face segmentation.

\myparagraph{Implementation Details. }Considering the inherent feature of 2D segmentation, if part $p$ of a face is invisible or occluded, it may lead to ${{\bm{C}_p} = \varnothing }$. In such a situation during training, we set $w_{prdl}^p = 0$ for these samples. We use Adam \cite{kingma2014adam} as the optimizer with an initial learning rate of $1e-4$. We use Farthest Point Sampling (FPS) \cite{moenning2003fast} to reduce the point number of ${V_{2d}^{skin}(\bm{\alpha} )}$ and $\bm{C}_{skin}$ to 3000, reducing computational consumption. Please refer to supplemental materials for more details.

\subsection{Metric}
In various VR/AR applications, 3DMMs are crucial in capturing facial motions or providing fine-grained regions covering facial features. One crucial objective in such applications is to ensure the alignment of overlapping facial parts between prediction and input. Widely used benchmarks \cite{sanyal2019learning,chai2022realy} typically rely on the 3D accuracy performance of reconstructions. However, there are instances where inconsistencies arise between 3D errors and 2D alignments. As shown in Fig.\ref{second}(b), comparing with 3DDFA-v2 \cite{guo2020towards}, DECA \cite{DECA:Siggraph2021} have better 2D eye region overlapping IoU (70.29\% \vs39.37\%) but a higher 3D forehead error ($1.88mm$ \vs$1.75mm$). To address this, we introduce Part IoU to emphasize the performance of overlap.

\myparagraph{Part IoU} is a new benchmark to quantify how well the part reconstruction ${V_{3d}^p(\bm{\alpha} )}$ aligns with their corresponding parts from the original face. The core idea is to measure the overlap of facial components between the reconstruction and the original image using IoU. The ground truth is a binary tensor $\{{\bm{M}_{p}}\}$ (as defined above). We render ${V_{3d}}(\bm{\alpha} )$ with a mean texture as an image, generate the predicted segmentation $\{\bm{M}_p^{pred}\}$ with \cite{Zheng2022DecoupledML}. The use of mean texture focuses the metric more on overlap effects than other factors, making it applicable to methods without texture-fitting \cite{feng2018joint, guo2020towards}. Part IoU $IoU_p$ of part $p$ can be obtained by: 
\begin{equation}
\begin{aligned}
\begin{array}{l}
IoU_p=IoU(\bm{M}_p^{pred},\bm{M}_p).
\end{array}
\end{aligned}
\end{equation}

MEAD \cite{kaisiyuan2020mead} is an emotional talking-face dataset. We test Part IoU by selecting $10$ individuals from MEAD, each contributing $50$ random different images. Part IoU measures the overlap performance between each part of the reconstruction and the ground truth. More detail is in the supplemental materials.

 \myparagraph{REALY}~\cite{chai2022realy} benchmark consists of $100$ scanned neutral expression faces, which are divided into four parts: nose, mouth, forehead (eyes and eyebrows), and cheek for 3D alignment and distance error calculation.

\subsection{Qualitative Comparison}
We conduct a comprehensive evaluation of our method with the state-of-the-art approaches, including PRNet \cite{feng2018joint}, MGCNet \cite{shang2020self}, Deep3D \cite{deng2019accurate}, 3DDFA-V2 \cite{guo2020towards}, HRN \cite{lei2023hierarchical} and DECA \cite{DECA:Siggraph2021}. The visualization of HRN and DECA uses the mid-frequency details and coarse shape (denoted as HRN-m and DECA-c) since their further steps only change the renderer's normal map, while no 3D refinement is made. As shown in Fig.~\ref{compare}, our results excel in capturing extreme expressions, even better than HRN-m which has fine reconstruction steps.

\subsection{Quantitative Comparison}
On both the Part IoU and REALY \cite{chai2022realy} benchmarks, our results outperforms the existing state-of-the-art methods. As shown in Tab.~\ref{part-iou-tab}, our method is almost always the highest overlap IoU across various facial parts with $73.82\%$ total average, demonstrating PRDL enhances the part alignment of reconstruction. PRDL also performs the best average 3D error on the REALY benchmark ($1.436mm$ in frontal-view and $1.442mm$ in side-view), as shown in Tab.~\ref{realy-tab}.

\begin{figure}[t]
\begin{center}
   \includegraphics[width=1\linewidth]{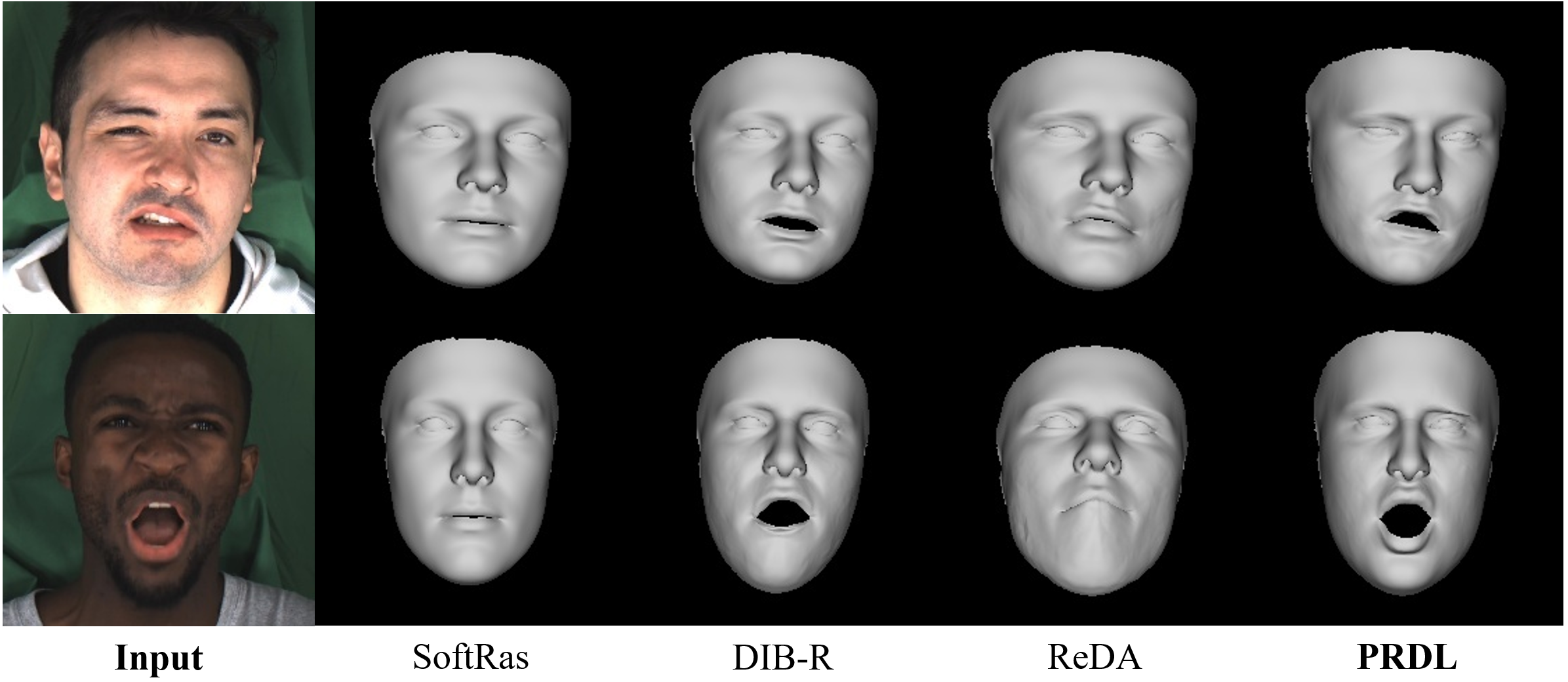}
\end{center}
 \vspace{-0.65cm}
   \caption{Comparison with the renderer-based geometric guidance of segmentation.
   }
\label{compare-render}
 \vspace{-0.55cm}
\end{figure}

\subsection{Ablation Study}
\label{abl}

\myparagraph{Ablation for PRDL and Synthetic Data.} We conduct quantitative ablation experiments for PRDL and synthetic data on REALY and Part IoU. As depicted in Table~\ref{part-iou-tab} and Table~\ref{realy-tab}, only introducing PRDL already yields superior results compared to all other methods (72.22\%, $1.468mm$, and $1.455mm$). Introducing synthetic data without PRDL demonstrates a significant improvement in Part IoU, but not as effectively as PRDL ({72.08\% \vs72.22\%}). Using both synthetic data and PRDL could lead to the best result.

\begin{figure}[t]
\begin{center}
   \includegraphics[width=1\linewidth]{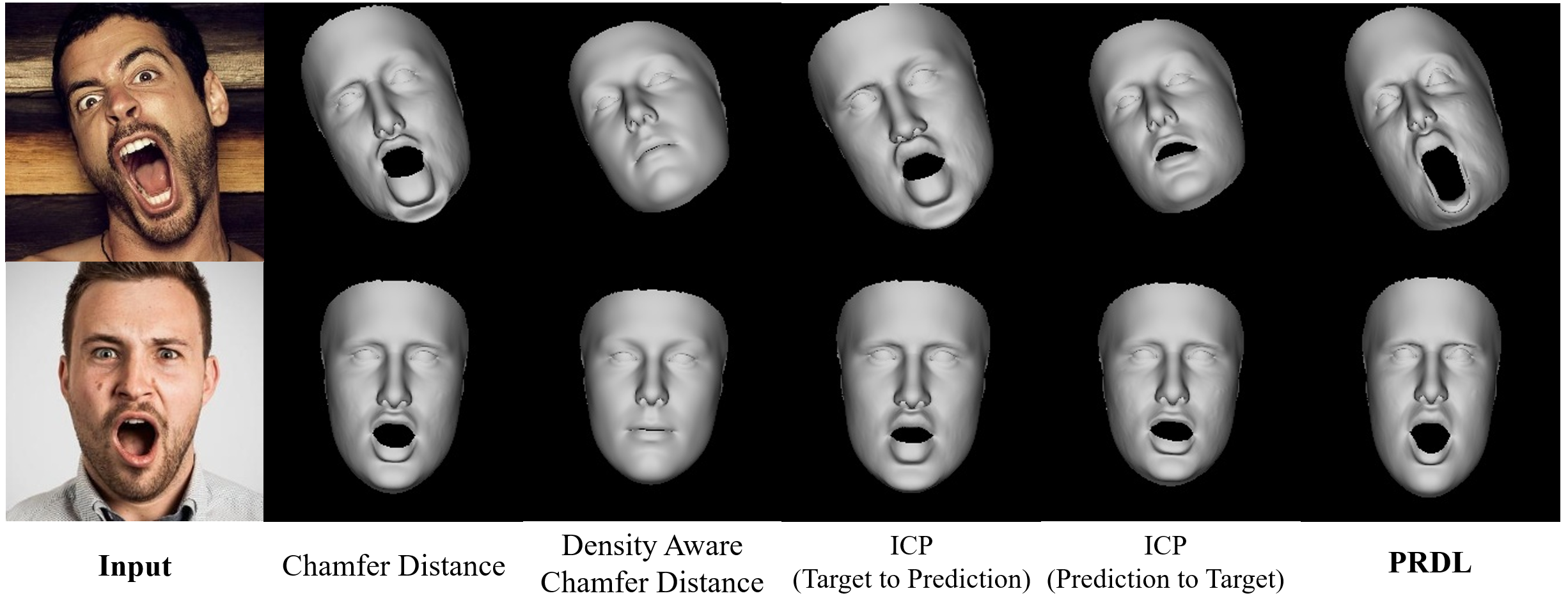}
\end{center}
 \vspace{-0.65cm}
   \caption{Comparison with the other point-driven-based geometric guidance of segmentation.
   }
\label{compare-coor}
 \vspace{-0.55cm}
\end{figure}

\myparagraph{Compare with the Differentiable Silhouette Renderers.} SoftRas \cite{liu2019soft} and DIB-R \cite{chen2019learning} are the two most widely used renderers, which serve as the basis for PyTorch3D \cite{ravi2020pytorch3d} and Kaolin \cite{KaolinLibrary}, respectively. Based on the image-fitting framework \cite{bfm_github}, we use them to render a silhouette of each face part and calculate the IoU loss with the ground truth. ReDA \cite{zhu2020reda} is also a renderer-based method using the geometric guidance of segmentation. Fig.\ref{compare-render} shows that PRDL is significantly better than these methods. It is essential to emphasize that all the results in Fig.\ref{compare-render} and Fig.\ref{compare-coor} do not include $\mathcal{L}_{{lmk}}$, $\mathcal{L}_{{pho}}$, and $\mathcal{L}_{{per}}$.

\myparagraph{Compare with the Other Point-Driven Optimization Methods.} One of the key insights of PRDL is transforming segmentation into points. Thus the 3DMM fitting becomes an optimization of two 2D point clouds until they share the same geometry. While an intuitive idea is incorporating the point-driven optimization methods like iterative closest points (ICP) \cite{icp4767965,icp121791,amberg2007optimal} or chamfer distance \cite{wu2021density}, these methods are predominantly rooted in nearest-neighbor principles, and solely opting for the minimum distance potentially leads to local optima. We compare PRDL with ICP \cite{yang2015go}, chamfer distance and density aware chamfer distance \cite{wu2021density} based on \cite{bfm_github}. Since the ICP distance can be calculated from target to prediction or vice versa, we provide both methods. As depicted in Fig.\ref{compare-coor}, PRDL outperforms other methods, producing outputs that align more accurately with the desired geometry. This superiority is attributed to the use of additional anchors and diverse statistical distances in PRDL. Referring to Fig.\ref{compare-render} and Fig.\ref{compare-coor}, PRDL stands out as the only loss capable of reconstructing effective results when the segmentation information is used independently. More comparison is in the supplemental materials.

\section{Conclusions}
This paper proposes a novel Part Re-projection Distance Loss (PRDL) to reconstruct 3D faces with the geometric guidance of facial part segmentation. Analysis proves that PRDL is superior to renderer-based and other point-driven optimization methods. We also provide a new emotional face expression dataset and a new 3D mesh part annotation to facilitate studies. Experiments further highlight the state-of-the-art performance of PRDL in achieving high-fidelity and better part alignment in 3D face reconstruction.

\section*{Acknowledgement}
This work was supported in part by Chinese National Natural Science Foundation Projects 62176256, U23B2054, 62276254, 62206280, the Beijing Science and Technology Plan Project Z231100005923033, Beijing Natural Science Foundation L221013, the Youth Innovation Promotion Association CAS Y2021131 and InnoHK program. 

% \balance
{
{\small
\bibliographystyle{ieeenat_fullname}
\bibliography{prdl_ref}

\begin{thebibliography}{60}
\providecommand{\natexlab}[1]{#1}
\providecommand{\url}[1]{\texttt{#1}}
\expandafter\ifx\csname urlstyle\endcsname\relax
  \providecommand{\doi}[1]{doi: #1}\else
  \providecommand{\doi}{doi: \begingroup \urlstyle{rm}\Url}\fi

\bibitem[bfm(2021)]{bfm_github}
3dmm model fitting using pytorch.
\newblock \url{https://github.com/ascust/3DMM-Fitting-Pytorch}, 2021.

\bibitem[Amberg et~al.(2007)Amberg, Romdhani, and Vetter]{amberg2007optimal}
Brian Amberg, Sami Romdhani, and Thomas Vetter.
\newblock Optimal step nonrigid icp algorithms for surface registration.
\newblock In \emph{2007 IEEE conference on computer vision and pattern recognition}, pages 1--8. IEEE, 2007.

\bibitem[Arun et~al.(1987)Arun, Huang, and Blostein]{icp4767965}
K.~S. Arun, T.~S. Huang, and S.~D. Blostein.
\newblock Least-squares fitting of two 3-d point sets.
\newblock \emph{IEEE Transactions on Pattern Analysis and Machine Intelligence}, PAMI-9\penalty0 (5):\penalty0 698--700, 1987.

\bibitem[Besl and McKay(1992)]{icp121791}
P.J. Besl and Neil~D. McKay.
\newblock A method for registration of 3-d shapes.
\newblock \emph{IEEE Transactions on Pattern Analysis and Machine Intelligence}, 14\penalty0 (2):\penalty0 239--256, 1992.

\bibitem[Blanz and Vetter(1999)]{blanz1999morphable}
Volker Blanz and Thomas Vetter.
\newblock A morphable model for the synthesis of 3d faces.
\newblock In \emph{Proceedings of the 26th annual conference on Computer graphics and interactive techniques}, pages 187--194, 1999.

\bibitem[Blanz and Vetter(2003)]{blanz2003face}
Volker Blanz and Thomas Vetter.
\newblock Face recognition based on fitting a 3d morphable model.
\newblock \emph{IEEE Transactions on pattern analysis and machine intelligence}, 25\penalty0 (9):\penalty0 1063--1074, 2003.

\bibitem[Chai et~al.(2022)Chai, Zhang, Ren, Kang, Xu, Zhe, Yuan, and Bao]{chai2022realy}
Zenghao Chai, Haoxian Zhang, Jing Ren, Di Kang, Zhengzhuo Xu, Xuefei Zhe, Chun Yuan, and Linchao Bao.
\newblock Realy: Rethinking the evaluation of 3d face reconstruction.
\newblock In \emph{Computer Vision--ECCV 2022: 17th European Conference, Tel Aviv, Israel, October 23--27, 2022, Proceedings, Part VIII}, pages 74--92. Springer, 2022.

\bibitem[Chen et~al.(2019)Chen, Ling, Gao, Smith, Lehtinen, Jacobson, and Fidler]{chen2019learning}
Wenzheng Chen, Huan Ling, Jun Gao, Edward Smith, Jaakko Lehtinen, Alec Jacobson, and Sanja Fidler.
\newblock Learning to predict 3d objects with an interpolation-based differentiable renderer.
\newblock \emph{Advances in neural information processing systems}, 32, 2019.

\bibitem[Deng et~al.(2019{\natexlab{a}})Deng, Guo, Xue, and Zafeiriou]{deng2019arcface}
Jiankang Deng, Jia Guo, Niannan Xue, and Stefanos Zafeiriou.
\newblock Arcface: Additive angular margin loss for deep face recognition.
\newblock In \emph{Proceedings of the IEEE Conference on Computer Vision and Pattern Recognition}, pages 4690--4699, 2019{\natexlab{a}}.

\bibitem[Deng et~al.(2020)Deng, Guo, Ververas, Kotsia, and Zafeiriou]{Deng2020CVPR}
Jiankang Deng, Jia Guo, Evangelos Ververas, Irene Kotsia, and Stefanos Zafeiriou.
\newblock Retinaface: Single-shot multi-level face localisation in the wild.
\newblock In \emph{CVPR}, 2020.

\bibitem[Deng et~al.(2019{\natexlab{b}})Deng, Yang, Xu, Chen, Jia, and Tong]{deng2019accurate}
Yu Deng, Jiaolong Yang, Sicheng Xu, Dong Chen, Yunde Jia, and Xin Tong.
\newblock Accurate 3d face reconstruction with weakly-supervised learning: From single image to image set.
\newblock In \emph{Proceedings of the IEEE/CVF conference on computer vision and pattern recognition workshops}, pages 0--0, 2019{\natexlab{b}}.

\bibitem[Egger et~al.(2018)Egger, Sch{\"o}nborn, Schneider, Kortylewski, Morel-Forster, Blumer, and Vetter]{egger2018occlusion}
Bernhard Egger, Sandro Sch{\"o}nborn, Andreas Schneider, Adam Kortylewski, Andreas Morel-Forster, Clemens Blumer, and Thomas Vetter.
\newblock Occlusion-aware 3d morphable models and an illumination prior for face image analysis.
\newblock \emph{International Journal of Computer Vision}, 126:\penalty0 1269--1287, 2018.

\bibitem[Feng et~al.(2018)Feng, Wu, Shao, Wang, and Zhou]{feng2018joint}
Yao Feng, Fan Wu, Xiaohu Shao, Yanfeng Wang, and Xi Zhou.
\newblock Joint 3d face reconstruction and dense alignment with position map regression network.
\newblock In \emph{Proceedings of the European conference on computer vision (ECCV)}, pages 534--551, 2018.

\bibitem[Feng et~al.(2021)Feng, Feng, Black, and Bolkart]{DECA:Siggraph2021}
Yao Feng, Haiwen Feng, Michael~J. Black, and Timo Bolkart.
\newblock Learning an animatable detailed {3D} face model from in-the-wild images.
\newblock 2021.

\bibitem[Fuji~Tsang et~al.(2022)Fuji~Tsang, Shugrina, Lafleche, Takikawa, Wang, Loop, Chen, Jatavallabhula, Smith, Rozantsev, Perel, Shen, Gao, Fidler, State, Gorski, Xiang, Li, Li, and Lebaredian]{KaolinLibrary}
Clement Fuji~Tsang, Maria Shugrina, Jean~Francois Lafleche, Towaki Takikawa, Jiehan Wang, Charles Loop, Wenzheng Chen, Krishna~Murthy Jatavallabhula, Edward Smith, Artem Rozantsev, Or Perel, Tianchang Shen, Jun Gao, Sanja Fidler, Gavriel State, Jason Gorski, Tommy Xiang, Jianing Li, Michael Li, and Rev Lebaredian.
\newblock Kaolin: A pytorch library for accelerating 3d deep learning research.
\newblock \url{https://github.com/NVIDIAGameWorks/kaolin}, 2022.

\bibitem[Genova et~al.(2018)Genova, Cole, Maschinot, Sarna, Vlasic, and Freeman]{genova2018unsupervised}
Kyle Genova, Forrester Cole, Aaron Maschinot, Aaron Sarna, Daniel Vlasic, and William~T Freeman.
\newblock Unsupervised training for 3d morphable model regression.
\newblock In \emph{Proceedings of the IEEE Conference on Computer Vision and Pattern Recognition}, pages 8377--8386, 2018.

\bibitem[Guo et~al.(2020)Guo, Zhu, Yang, Yang, Lei, and Li]{guo2020towards}
Jianzhu Guo, Xiangyu Zhu, Yang Yang, Fan Yang, Zhen Lei, and Stan~Z Li.
\newblock Towards fast, accurate and stable 3d dense face alignment.
\newblock pages 152--168, 2020.

\bibitem[He et~al.(2016)He, Zhang, Ren, and Sun]{he2016deep}
Kaiming He, Xiangyu Zhang, Shaoqing Ren, and Jian Sun.
\newblock Deep residual learning for image recognition.
\newblock In \emph{Proceedings of the IEEE conference on computer vision and pattern recognition}, pages 770--778, 2016.

\bibitem[Kao et~al.(2023)Kao, Pan, Xu, Lyu, Zhu, Chang, Li, and Lei]{xu10127617}
Yueying Kao, Bowen Pan, Miao Xu, Jiangjing Lyu, Xiangyu Zhu, Yuanzhang Chang, Xiaobo Li, and Zhen Lei.
\newblock Toward 3d face reconstruction in perspective projection: Estimating 6dof face pose from monocular image.
\newblock \emph{IEEE Transactions on Image Processing}, 32:\penalty0 3080--3091, 2023.

\bibitem[Kartynnik et~al.(2019)Kartynnik, Ablavatski, Grishchenko, and Grundmann]{kartynnik2019real}
Yury Kartynnik, Artsiom Ablavatski, Ivan Grishchenko, and Matthias Grundmann.
\newblock Real-time facial surface geometry from monocular video on mobile gpus.
\newblock \emph{arXiv preprint arXiv:1907.06724}, 2019.

\bibitem[Kato et~al.(2018)Kato, Ushiku, and Harada]{kato2018neural}
Hiroharu Kato, Yoshitaka Ushiku, and Tatsuya Harada.
\newblock Neural 3d mesh renderer.
\newblock In \emph{Proceedings of the IEEE conference on computer vision and pattern recognition}, pages 3907--3916, 2018.

\bibitem[Kato et~al.(2020)Kato, Beker, Morariu, Ando, Matsuoka, Kehl, and Gaidon]{kato2020differentiable}
Hiroharu Kato, Deniz Beker, Mihai Morariu, Takahiro Ando, Toru Matsuoka, Wadim Kehl, and Adrien Gaidon.
\newblock Differentiable rendering: A survey.
\newblock \emph{arXiv preprint arXiv:2006.12057}, 2020.

\bibitem[Kingma and Ba(2014)]{kingma2014adam}
Diederik~P Kingma and Jimmy Ba.
\newblock Adam: A method for stochastic optimization.
\newblock \emph{arXiv preprint arXiv:1412.6980}, 2014.

\bibitem[Lee et~al.(2020)Lee, Liu, Wu, and Luo]{CelebAMask-HQ}
Cheng-Han Lee, Ziwei Liu, Lingyun Wu, and Ping Luo.
\newblock Maskgan: Towards diverse and interactive facial image manipulation.
\newblock In \emph{IEEE Conference on Computer Vision and Pattern Recognition (CVPR)}, 2020.

\bibitem[Lei et~al.(2023)Lei, Ren, Feng, Cui, and Xie]{lei2023hierarchical}
Biwen Lei, Jianqiang Ren, Mengyang Feng, Miaomiao Cui, and Xuansong Xie.
\newblock A hierarchical representation network for accurate and detailed face reconstruction from in-the-wild images.
\newblock In \emph{Proceedings of the IEEE/CVF Conference on Computer Vision and Pattern Recognition}, pages 394--403, 2023.

\bibitem[Li et~al.(2021)Li, Morel-Forster, Vetter, Egger, and Kortylewski]{li2021fit}
Chunlu Li, Andreas Morel-Forster, Thomas Vetter, Bernhard Egger, and Adam Kortylewski.
\newblock To fit or not to fit: Model-based face reconstruction and occlusion segmentation from weak supervision.
\newblock \emph{arXiv preprint arXiv:2106.09614}, 2021.

\bibitem[Li et~al.(2020)Li, Bladin, Zhao, Chinara, Ingraham, Xiang, Ren, Prasad, Kishore, Xing, and Li]{li2020learning}
Ruilong Li, Karl Bladin, Yajie Zhao, Chinmay Chinara, Owen Ingraham, Pengda Xiang, Xinglei Ren, Pratusha Prasad, Bipin Kishore, Jun Xing, and Hao Li.
\newblock Learning formation of physically-based face attributes.
\newblock 2020.

\bibitem[Li and Deng(2019{\natexlab{a}})]{DBLP:journals/ijcv/ShangD19}
Shan Li and Weihong Deng.
\newblock Blended emotion in-the-wild: Multi-label facial expression recognition using crowdsourced annotations and deep locality feature learning.
\newblock \emph{International Journal of Computer Vision}, 127\penalty0 (6-7):\penalty0 884--906, 2019{\natexlab{a}}.

\bibitem[Li and Deng(2019{\natexlab{b}})]{li2019reliable}
Shan Li and Weihong Deng.
\newblock Reliable crowdsourcing and deep locality-preserving learning for unconstrained facial expression recognition.
\newblock \emph{IEEE Transactions on Image Processing}, 28\penalty0 (1):\penalty0 356--370, 2019{\natexlab{b}}.

\bibitem[Li et~al.(2017)Li, Bolkart, Black, Li, and Romero]{FLAME:SiggraphAsia2017}
Tianye Li, Timo Bolkart, Michael.~J. Black, Hao Li, and Javier Romero.
\newblock Learning a model of facial shape and expression from {4D} scans.
\newblock \emph{ACM Transactions on Graphics, (Proc. SIGGRAPH Asia)}, 36\penalty0 (6):\penalty0 194:1--194:17, 2017.

\bibitem[Lin et~al.(2019)Lin, Yang, Chen, Zeng, Wen, and Yuan]{lin2019face}
Jinpeng Lin, Hao Yang, Dong Chen, Ming Zeng, Fang Wen, and Lu Yuan.
\newblock Face parsing with roi tanh-warping.
\newblock In \emph{Proceedings of the IEEE/CVF Conference on Computer Vision and Pattern Recognition}, pages 5654--5663, 2019.

\bibitem[Lin et~al.(2021)Lin, Shen, Wang, and Pantic]{lin2021roi}
Yiming Lin, Jie Shen, Yujiang Wang, and Maja Pantic.
\newblock Roi tanh-polar transformer network for face parsing in the wild.
\newblock \emph{Image and Vision Computing}, 112:\penalty0 104190, 2021.

\bibitem[Liu et~al.(2019)Liu, Li, Chen, and Li]{liu2019soft}
Shichen Liu, Tianye Li, Weikai Chen, and Hao Li.
\newblock Soft rasterizer: A differentiable renderer for image-based 3d reasoning.
\newblock In \emph{Proceedings of the IEEE/CVF International Conference on Computer Vision}, pages 7708--7717, 2019.

\bibitem[Liu et~al.(2020)Liu, Shi, Shen, Si, Wang, and Mei]{liu2020new}
Yinglu Liu, Hailin Shi, Hao Shen, Yue Si, Xiaobo Wang, and Tao Mei.
\newblock A new dataset and boundary-attention semantic segmentation for face parsing.
\newblock In \emph{AAAI}, pages 11637--11644, 2020.

\bibitem[Liu et~al.(2015)Liu, Luo, Wang, and Tang]{liu2015faceattributes}
Ziwei Liu, Ping Luo, Xiaogang Wang, and Xiaoou Tang.
\newblock Deep learning face attributes in the wild.
\newblock In \emph{Proceedings of International Conference on Computer Vision (ICCV)}, 2015.

\bibitem[Loper and Black(2014)]{loper2014opendr}
Matthew~M Loper and Michael~J Black.
\newblock Opendr: An approximate differentiable renderer.
\newblock In \emph{Computer Vision--ECCV 2014: 13th European Conference, Zurich, Switzerland, September 6-12, 2014, Proceedings, Part VII 13}, pages 154--169. Springer, 2014.

\bibitem[Martyniuk et~al.(2022)Martyniuk, Kupyn, Kurlyak, Krashenyi, Matas, and Sharmanska]{dad3dheads}
Tetiana Martyniuk, Orest Kupyn, Yana Kurlyak, Igor Krashenyi, Ji\v{r}i Matas, and Viktoriia Sharmanska.
\newblock Dad-3dheads: A large-scale dense, accurate and diverse dataset for 3d head alignment from a single image.
\newblock In \emph{Proc. IEEE Conf. on Computer Vision and Pattern Recognition (CVPR)}, 2022.

\bibitem[Moenning and Dodgson(2003)]{moenning2003fast}
Carsten Moenning and Neil~A Dodgson.
\newblock Fast marching farthest point sampling.
\newblock Technical report, University of Cambridge, Computer Laboratory, 2003.

\bibitem[Paszke et~al.(2019)Paszke, Gross, Massa, Lerer, Bradbury, Chanan, Killeen, Lin, Gimelshein, Antiga, et~al.]{paszke2019pytorch}
Adam Paszke, Sam Gross, Francisco Massa, Adam Lerer, James Bradbury, Gregory Chanan, Trevor Killeen, Zeming Lin, Natalia Gimelshein, Luca Antiga, et~al.
\newblock Pytorch: An imperative style, high-performance deep learning library.
\newblock \emph{Advances in neural information processing systems}, 32, 2019.

\bibitem[Ramamoorthi and Hanrahan(2001)]{ramamoorthi2001efficient}
Ravi Ramamoorthi and Pat Hanrahan.
\newblock An efficient representation for irradiance environment maps.
\newblock In \emph{Proceedings of the 28th annual conference on Computer graphics and interactive techniques}, pages 497--500, 2001.

\bibitem[Raman et~al.(2023)Raman, Hewitt, Wood, and Baltru{\v{s}}aitis]{raman2023mesh}
Chirag Raman, Charlie Hewitt, Erroll Wood, and Tadas Baltru{\v{s}}aitis.
\newblock Mesh-tension driven expression-based wrinkles for synthetic faces.
\newblock In \emph{Proceedings of the IEEE/CVF Winter Conference on Applications of Computer Vision}, pages 3515--3525, 2023.

\bibitem[Ravi et~al.(2020)Ravi, Reizenstein, Novotny, Gordon, Lo, Johnson, and Gkioxari]{ravi2020pytorch3d}
Nikhila Ravi, Jeremy Reizenstein, David Novotny, Taylor Gordon, Wan-Yen Lo, Justin Johnson, and Georgia Gkioxari.
\newblock Accelerating 3d deep learning with pytorch3d.
\newblock \emph{arXiv:2007.08501}, 2020.

\bibitem[Sagonas et~al.(2013)Sagonas, Tzimiropoulos, Zafeiriou, and Pantic]{sagonas2013300}
Christos Sagonas, Georgios Tzimiropoulos, Stefanos Zafeiriou, and Maja Pantic.
\newblock 300 faces in-the-wild challenge: The first facial landmark localization challenge.
\newblock In \emph{Proceedings of the IEEE international conference on computer vision workshops}, pages 397--403, 2013.

\bibitem[Sanyal et~al.(2019)Sanyal, Bolkart, Feng, and Black]{sanyal2019learning}
Soubhik Sanyal, Timo Bolkart, Haiwen Feng, and Michael~J Black.
\newblock Learning to regress 3d face shape and expression from an image without 3d supervision.
\newblock In \emph{Proceedings of the IEEE/CVF Conference on Computer Vision and Pattern Recognition}, pages 7763--7772, 2019.

\bibitem[Shang et~al.(2020)Shang, Shen, Li, Zhou, Zhen, Fang, and Quan]{shang2020self}
Jiaxiang Shang, Tianwei Shen, Shiwei Li, Lei Zhou, Mingmin Zhen, Tian Fang, and Long Quan.
\newblock Self-supervised monocular 3d face reconstruction by occlusion-aware multi-view geometry consistency.
\newblock In \emph{Computer Vision--ECCV 2020: 16th European Conference, Glasgow, UK, August 23--28, 2020, Proceedings, Part XV}, pages 53--70. Springer, 2020.

\bibitem[Shreiner et~al.(2009)Shreiner, Group, et~al.]{shreiner2009opengl}
Dave Shreiner, Bill The Khronos OpenGL ARB~Working Group, et~al.
\newblock \emph{OpenGL programming guide: the official guide to learning OpenGL, versions 3.0 and 3.1}.
\newblock Pearson Education, 2009.

\bibitem[Sun et~al.(2022)Sun, Wang, Shi, Wang, Wang, and Liu]{sun2022ide}
Jingxiang Sun, Xuan Wang, Yichun Shi, Lizhen Wang, Jue Wang, and Yebin Liu.
\newblock Ide-3d: Interactive disentangled editing for high-resolution 3d-aware portrait synthesis.
\newblock \emph{ACM Transactions on Graphics (TOG)}, 41\penalty0 (6):\penalty0 1--10, 2022.

\bibitem[Tewari et~al.(2021)Tewari, Seidel, Elgharib, Theobalt, et~al.]{tewari2021learning}
Ayush Tewari, Hans-Peter Seidel, Mohamed Elgharib, Christian Theobalt, et~al.
\newblock Learning complete 3d morphable face models from images and videos.
\newblock In \emph{Proceedings of the IEEE/CVF Conference on Computer Vision and Pattern Recognition}, pages 3361--3371, 2021.

\bibitem[University~of Basel and Research(2017)]{bfm_seg_github}
Graphics University~of Basel and Vision Research.
\newblock parametric-face-image-generator.
\newblock \url{https://github.com/unibas-gravis/parametric-face-image-generator}, 2017.

\bibitem[Wang et~al.(2020)Wang, Wu, Song, Yang, Wu, Qian, He, Qiao, and Loy]{kaisiyuan2020mead}
Kaisiyuan Wang, Qianyi Wu, Linsen Song, Zhuoqian Yang, Wayne Wu, Chen Qian, Ran He, Yu Qiao, and Chen~Change Loy.
\newblock Mead: A large-scale audio-visual dataset for emotional talking-face generation.
\newblock In \emph{ECCV}, 2020.

\bibitem[Wang et~al.(2022)Wang, Chen, Yu, Ma, Li, and Liu]{wang2022faceverse}
Lizhen Wang, Zhiyuan Chen, Tao Yu, Chenguang Ma, Liang Li, and Yebin Liu.
\newblock Faceverse: a fine-grained and detail-controllable 3d face morphable model from a hybrid dataset.
\newblock In \emph{Proceedings of the IEEE/CVF Conference on Computer Vision and Pattern Recognition}, pages 20333--20342, 2022.

\bibitem[Wood et~al.(2021)Wood, Baltru{\v{s}}aitis, Hewitt, Dziadzio, Cashman, and Shotton]{wood2021fake}
Erroll Wood, Tadas Baltru{\v{s}}aitis, Charlie Hewitt, Sebastian Dziadzio, Thomas~J Cashman, and Jamie Shotton.
\newblock Fake it till you make it: face analysis in the wild using synthetic data alone.
\newblock In \emph{Proceedings of the IEEE/CVF international conference on computer vision}, pages 3681--3691, 2021.

\bibitem[Wu et~al.(2021)Wu, Pan, Zhang, Wang, Liu, and Lin]{wu2021density}
Tong Wu, Liang Pan, Junzhe Zhang, Tai Wang, Ziwei Liu, and Dahua Lin.
\newblock Density-aware chamfer distance as a comprehensive metric for point cloud completion.
\newblock \emph{arXiv preprint arXiv:2111.12702}, 2021.

\bibitem[Yang et~al.(2015)Yang, Li, Campbell, and Jia]{yang2015go}
Jiaolong Yang, Hongdong Li, Dylan Campbell, and Yunde Jia.
\newblock Go-icp: A globally optimal solution to 3d icp point-set registration.
\newblock \emph{IEEE transactions on pattern analysis and machine intelligence}, 38\penalty0 (11):\penalty0 2241--2254, 2015.

\bibitem[Zheng et~al.(2022)Zheng, Deng, Zhu, Li, and Zafeiriou]{Zheng2022DecoupledML}
Qi Zheng, Jiankang Deng, Zheng Zhu, Ying Li, and Stefanos Zafeiriou.
\newblock Decoupled multi-task learning with cyclical self-regulation for face parsing.
\newblock In \emph{Computer Vision and Pattern Recognition}, 2022.

\bibitem[Zhu et~al.(2020)Zhu, Wu, Chen, Vesdapunt, and Wang]{zhu2020reda}
Wenbin Zhu, HsiangTao Wu, Zeyu Chen, Noranart Vesdapunt, and Baoyuan Wang.
\newblock Reda: reinforced differentiable attribute for 3d face reconstruction.
\newblock In \emph{Proceedings of the IEEE/CVF Conference on Computer Vision and Pattern Recognition}, pages 4958--4967, 2020.

\bibitem[Zhu et~al.(2015)Zhu, Lei, Yan, Yi, and Li]{zhu2015high}
Xiangyu Zhu, Zhen Lei, Junjie Yan, Dong Yi, and Stan~Z Li.
\newblock High-fidelity pose and expression normalization for face recognition in the wild.
\newblock In \emph{Proceedings of the IEEE conference on computer vision and pattern recognition}, pages 787--796, 2015.

\bibitem[Zhu et~al.(2017)Zhu, Liu, Lei, and Li]{zhu2017face}
Xiangyu Zhu, Xiaoming Liu, Zhen Lei, and Stan~Z Li.
\newblock Face alignment in full pose range: A 3d total solution.
\newblock \emph{IEEE transactions on pattern analysis and machine intelligence}, 41\penalty0 (1):\penalty0 78--92, 2017.

\bibitem[Zhu et~al.(2022)Zhu, Yu, Huang, Lei, Wang, and Li]{zhu2022beyond}
Xiangyu Zhu, Chang Yu, Di Huang, Zhen Lei, Hao Wang, and Stan~Z Li.
\newblock Beyond 3dmm: Learning to capture high-fidelity 3d face shape.
\newblock \emph{IEEE Transactions on Pattern Analysis and Machine Intelligence}, 2022.

\bibitem[Zielonka et~al.(2022)Zielonka, Bolkart, and Thies]{zielonka2022towards}
Wojciech Zielonka, Timo Bolkart, and Justus Thies.
\newblock Towards metrical reconstruction of human faces.
\newblock In \emph{European Conference on Computer Vision}, pages 250--269. Springer, 2022.

\end{thebibliography}


\begin{thebibliography}{15}
\providecommand{\natexlab}[1]{#1}
\providecommand{\url}[1]{\texttt{#1}}
\expandafter\ifx\csname urlstyle\endcsname\relax
  \providecommand{\doi}[1]{doi: #1}\else
  \providecommand{\doi}{doi: \begingroup \urlstyle{rm}\Url}\fi

\bibitem[bfm(2021)]{bfm_githubsupp}
3dmm model fitting using pytorch.
\newblock \url{https://github.com/ascust/3DMM-Fitting-Pytorch}, 2021.

\bibitem[Bradski(2000)]{bradski2000opencvsupp}
Gary Bradski.
\newblock The opencv library.
\newblock \emph{Dr. Dobb's Journal: Software Tools for the Professional Programmer}, 25\penalty0 (11):\penalty0 120--123, 2000.

\bibitem[Deng et~al.(2019)Deng, Yang, Xu, Chen, Jia, and Tong]{deng2019accuratesupp}
Yu Deng, Jiaolong Yang, Sicheng Xu, Dong Chen, Yunde Jia, and Xin Tong.
\newblock Accurate 3d face reconstruction with weakly-supervised learning: From single image to image set.
\newblock In \emph{Proceedings of the IEEE/CVF conference on computer vision and pattern recognition workshops}, pages 0--0, 2019.

\bibitem[Feng et~al.(2018)Feng, Wu, Shao, Wang, and Zhou]{feng2018jointsupp}
Yao Feng, Fan Wu, Xiaohu Shao, Yanfeng Wang, and Xi Zhou.
\newblock Joint 3d face reconstruction and dense alignment with position map regression network.
\newblock In \emph{Proceedings of the European conference on computer vision (ECCV)}, pages 534--551, 2018.

\bibitem[Feng et~al.(2021)Feng, Feng, Black, and Bolkart]{DECA:Siggraph2021supp}
Yao Feng, Haiwen Feng, Michael~J. Black, and Timo Bolkart.
\newblock Learning an animatable detailed {3D} face model from in-the-wild images.
\newblock 2021.

\bibitem[Guo et~al.(2020)Guo, Zhu, Yang, Yang, Lei, and Li]{guo2020towardssupp}
Jianzhu Guo, Xiangyu Zhu, Yang Yang, Fan Yang, Zhen Lei, and Stan~Z Li.
\newblock Towards fast, accurate and stable 3d dense face alignment.
\newblock pages 152--168, 2020.

\bibitem[Le et~al.(2012)Le, Brandt, Lin, Bourdev, and Huang]{Le2012InteractiveFFsupp}
Vuong Le, Jonathan Brandt, Zhe~L. Lin, Lubomir~D. Bourdev, and Thomas~S. Huang.
\newblock Interactive facial feature localization.
\newblock In \emph{European Conference on Computer Vision}, 2012.

\bibitem[Lee et~al.(2020)Lee, Liu, Wu, and Luo]{CelebAMask-HQsupp}
Cheng-Han Lee, Ziwei Liu, Lingyun Wu, and Ping Luo.
\newblock Maskgan: Towards diverse and interactive facial image manipulation.
\newblock In \emph{IEEE Conference on Computer Vision and Pattern Recognition (CVPR)}, 2020.

\bibitem[Lei et~al.(2023)Lei, Ren, Feng, Cui, and Xie]{lei2023hierarchicalsupp}
Biwen Lei, Jianqiang Ren, Mengyang Feng, Miaomiao Cui, and Xuansong Xie.
\newblock A hierarchical representation network for accurate and detailed face reconstruction from in-the-wild images.
\newblock In \emph{Proceedings of the IEEE/CVF Conference on Computer Vision and Pattern Recognition}, pages 394--403, 2023.

\bibitem[Lin et~al.(2021)Lin, Shen, Wang, and Pantic]{lin2021roisupp}
Yiming Lin, Jie Shen, Yujiang Wang, and Maja Pantic.
\newblock Roi tanh-polar transformer network for face parsing in the wild.
\newblock \emph{Image and Vision Computing}, 112:\penalty0 104190, 2021.

\bibitem[Paysan et~al.(2009)Paysan, Knothe, Amberg, Romdhani, and Vetter]{paysan20093dsupp}
Pascal Paysan, Reinhard Knothe, Brian Amberg, Sami Romdhani, and Thomas Vetter.
\newblock A 3d face model for pose and illumination invariant face recognition.
\newblock In \emph{2009 sixth IEEE international conference on advanced video and signal based surveillance}, pages 296--301. Ieee, 2009.

\bibitem[Shang et~al.(2020)Shang, Shen, Li, Zhou, Zhen, Fang, and Quan]{shang2020selfsupp}
Jiaxiang Shang, Tianwei Shen, Shiwei Li, Lei Zhou, Mingmin Zhen, Tian Fang, and Long Quan.
\newblock Self-supervised monocular 3d face reconstruction by occlusion-aware multi-view geometry consistency.
\newblock In \emph{Computer Vision--ECCV 2020: 16th European Conference, Glasgow, UK, August 23--28, 2020, Proceedings, Part XV}, pages 53--70. Springer, 2020.

\bibitem[Wang et~al.(2020)Wang, Wu, Song, Yang, Wu, Qian, He, Qiao, and Loy]{kaisiyuan2020meadsupp}
Kaisiyuan Wang, Qianyi Wu, Linsen Song, Zhuoqian Yang, Wayne Wu, Chen Qian, Ran He, Yu Qiao, and Chen~Change Loy.
\newblock Mead: A large-scale audio-visual dataset for emotional talking-face generation.
\newblock In \emph{ECCV}, 2020.

\bibitem[Wang et~al.(2022)Wang, Chen, Yu, Ma, Li, and Liu]{wang2022faceversesupp}
Lizhen Wang, Zhiyuan Chen, Tao Yu, Chenguang Ma, Liang Li, and Yebin Liu.
\newblock Faceverse: a fine-grained and detail-controllable 3d face morphable model from a hybrid dataset.
\newblock In \emph{Proceedings of the IEEE/CVF Conference on Computer Vision and Pattern Recognition}, pages 20333--20342, 2022.

\bibitem[Zheng et~al.(2022)Zheng, Deng, Zhu, Li, and Zafeiriou]{Zheng2022DecoupledMLsupp}
Qi Zheng, Jiankang Deng, Zheng Zhu, Ying Li, and Stefanos Zafeiriou.
\newblock Decoupled multi-task learning with cyclical self-regulation for face parsing.
\newblock In \emph{Computer Vision and Pattern Recognition}, 2022.

\end{thebibliography}
}
}

\newpage

\appendix

\maketitlesupplementary

\section{More Analysis about PRDL}
\begin{figure}[t]
\begin{center}
\includegraphics[width=1\linewidth]{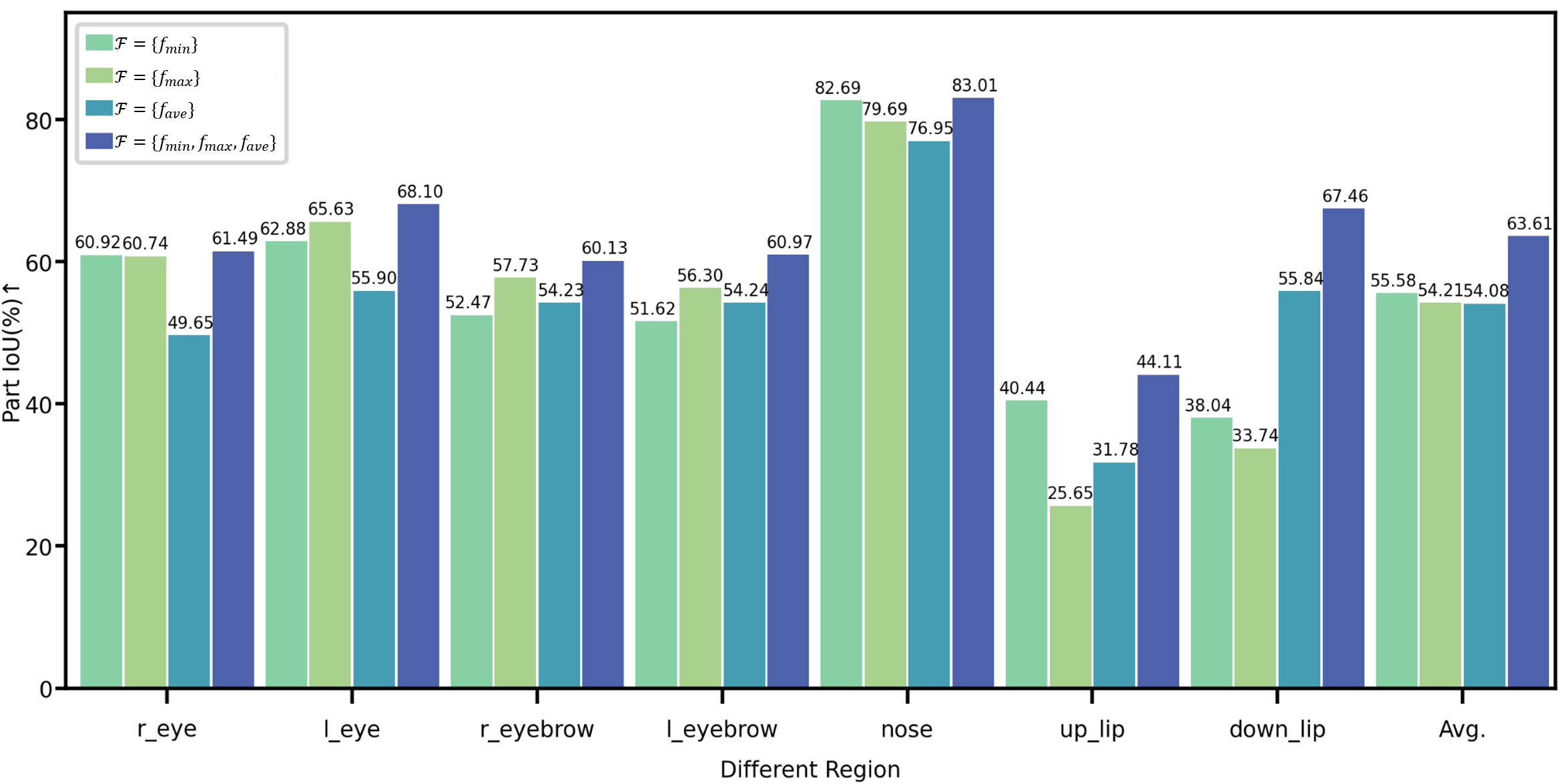 }
\end{center}
\vspace{-0.6cm}
\caption{Quantitative comparison on Part IoU benchmark for $f_{min},f_{max}$, and $f_{ave}$. 
   }
\label{sup-grad-iou}
\vspace{-0.2cm}
\end{figure}

\myparagraph{Ablation Study for $\bm{f_{min}}$, $\bm{ f_{max}}$, and $\bm{f_{ave}}$.} In the main paper, we have extensively analyzed the gradient of PRDL in the case of $f_{min}$. In the supplementary material, we leverage the image-fitting framework \citesupp{bfm_githubsupp} to further elucidate the roles of $f_{min}$, $f_{max}$ and $f_{ave}$ based on Part IoU benchmark. As depicted in Fig.~\ref{sup-grad-iou}, individually applying $f_{min}$, $f_{max}$, or $f_{ave}$ yields satisfactory results, and their combined application leads to a significant improvement (63.61\% in average). It should be noted that all results in Fig.~\ref{sup-grad-iou} do not include $\mathcal{L}_{{lmk}}$, $\mathcal{L}_{{pho}}$, and $\mathcal{L}_{{per}}$. 

\begin{figure}[t]
\begin{center}
\includegraphics[width=1\linewidth]{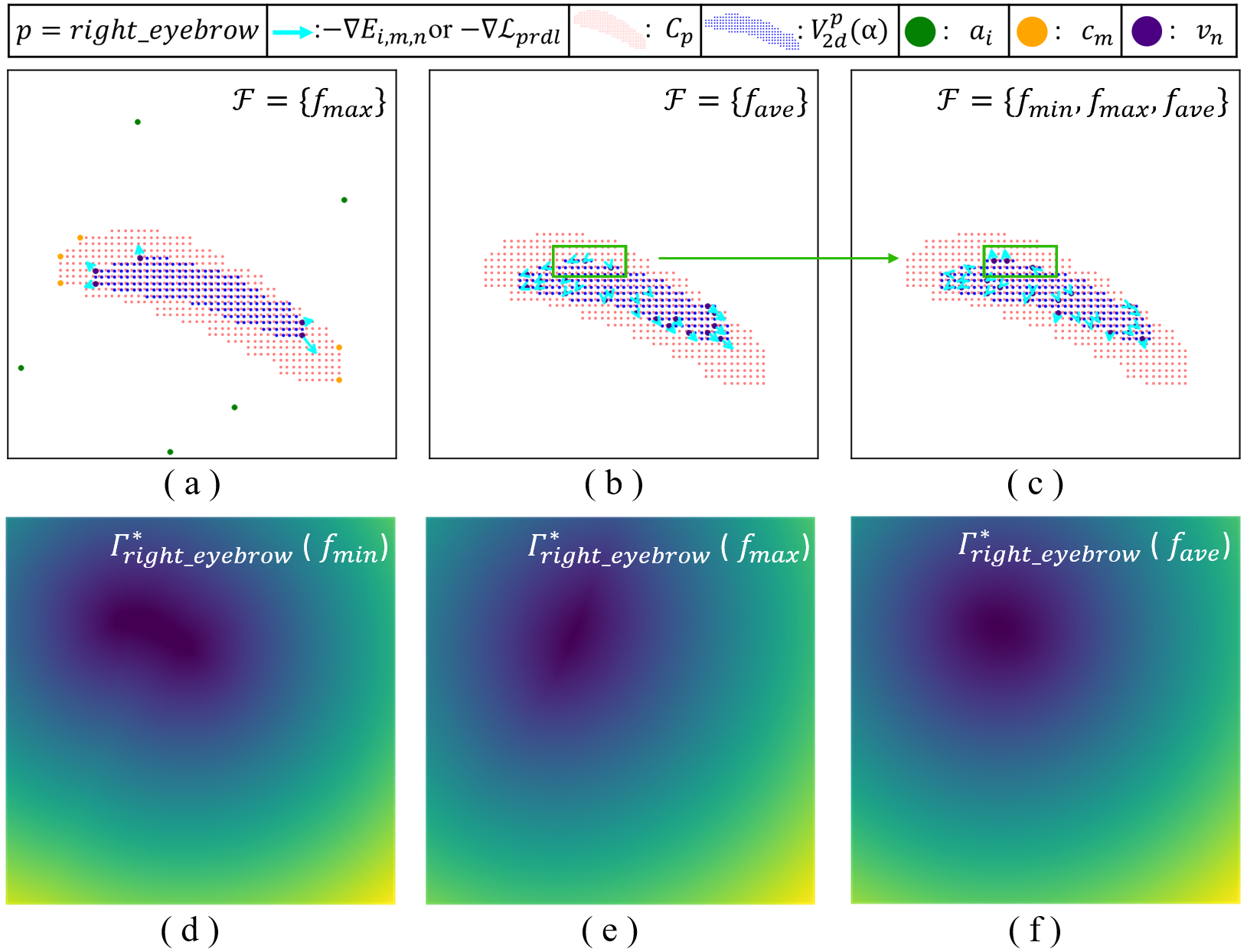 }
\end{center}
\vspace{-0.6cm}
\caption{More analysis about PRDL when $p=$ right\_eyebrow. (a) Visualization of $-\nabla {E_{i,m,n}}$ when $ \bm{\mathcal{F}} = \{f_{max}\}$. (b) and (c) depict the visualizations of $-\nabla {\mathcal{L}_{prdl}}$ when $ \bm{\mathcal{F}} = \{f_{ave}\}$ and $ \bm{\mathcal{F}} = \{f_{min},f_{max},f_{ave}\}$, respectively. (d), (e), and (f) visualize $\bm{\Gamma}_p^*$ in three channels ($ f_{min},f_{max}$, and $f_{ave}$).
   }
\vspace{-0.4cm}
\label{sup-grad}
\end{figure}

\myparagraph{More Gradient Analysis about PRDL.} The above results indicate that the adoption of various distance measures is beneficial, which are also demonstrated in Fig.~\ref{sup-grad}. We select a subset of $v_{n}$ for gradient analysis. The effect of $f_{max}$ is similar to that of $f_{min}$, with the only difference being the selection of points. $f_{ave}$ influences the entire ${V_{2d}^p(\bm{\alpha} )}$. As indicated by the green box in Fig.~\ref{sup-grad}(b) and Fig.~\ref{sup-grad}(c), the gradient of $ \bm{\mathcal{F}} = \{f_{min},f_{max},f_{ave}\}$ is the most effective as it can correct some errors that arise when $f_{ave}$ acts alone in certain scenarios. Aqua-colored arrows are used to indicate the direction of the gradient, with the length serving merely as an illustration. Reshape $\bm{\Gamma}_p^*$ (${\mathbb{R}^{|\bm{A}| \times |\bm{\mathcal{F}}|}} \to  \mathbb{R}^{H\times W \times |\bm{\mathcal{F}}|}$) to visualize its three channels ($f_{min},f_{max},f_{ave}$) separately in Fig.~\ref{sup-grad}(d), Fig.~\ref{sup-grad}(e), and Fig.~\ref{sup-grad}(f). Compared to the regression target of the renderer-based Loss, $\bm{\Gamma}_p^*$ in PRDL is more informative and more conducive for fitting.

\section{More Implementation Details}

\begin{figure}[t]
\begin{center}
\includegraphics[width=1\linewidth]{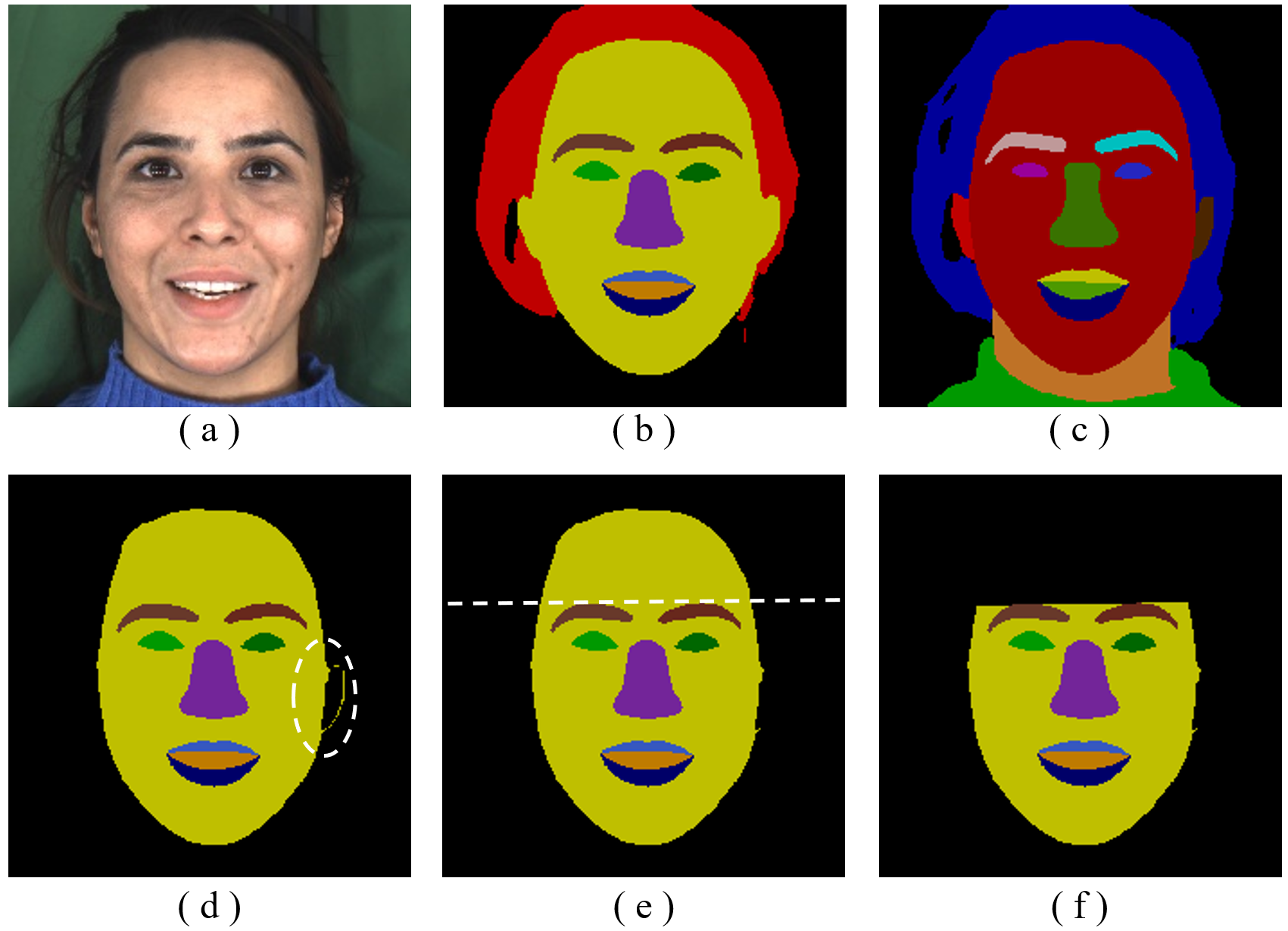 }
\end{center}
\vspace{-0.6cm}

\caption{Remove the ear, filter noise and dynamically remove the forehead region according to the position of the eyebrows.
   }
\vspace{-0.2cm}
\label{sup-seg}
\end{figure}

\myparagraph{Transforming Segmentation to 2D Points.} Two widely recognized definitions of 2D face segmentation regions are Helen \citesupp{Le2012InteractiveFFsupp} or iBugMask \citesupp{lin2021roisupp} and CelebAMask-HQ \citesupp{CelebAMask-HQsupp}, which divide the face and related areas into 11 parts and 19 parts, respectively. As shown in Fig.~\ref{sup-seg}, we employ the state-of-the-art method DML-CSR \citesupp{Zheng2022DecoupledMLsupp} for face segmentation. The results of the above two segmentation definitions are shown in Fig.~\ref{sup-seg}(b) and Fig.~\ref{sup-seg}(c), respectively. Through practical experimentation, we find that the 11-part method yields more accurate results. However, the segmentation of the ear regions from this method does not align well with the face model and needs to be removed. Consequently, we remove the corresponding ear regions from Fig.~\ref{sup-seg}(b) based on Fig.~\ref{sup-seg}(c), resulting in Fig.~\ref{sup-seg}(d). Typically, Fig.~\ref{sup-seg}(d) contains noise as indicated by the white dashed circle. To handle this, we identify the noise \citesupp{bradski2000opencvsupp} and eliminate these isolated regions, yielding the outcome depicted in Fig.~\ref{sup-seg}(e). To mitigate the impact of the region above the eyebrows, which is often obscured by hair, we transformed the eyebrows into 2D coordinates, identified their tangents (represented by white dashed lines in Fig.~\ref{sup-seg}(e)), and dynamically removed the area above the eyebrows. The final result is presented in Fig.~\ref{sup-seg}(f).

\begin{figure}[t]
\begin{center}
\includegraphics[width=1\linewidth]{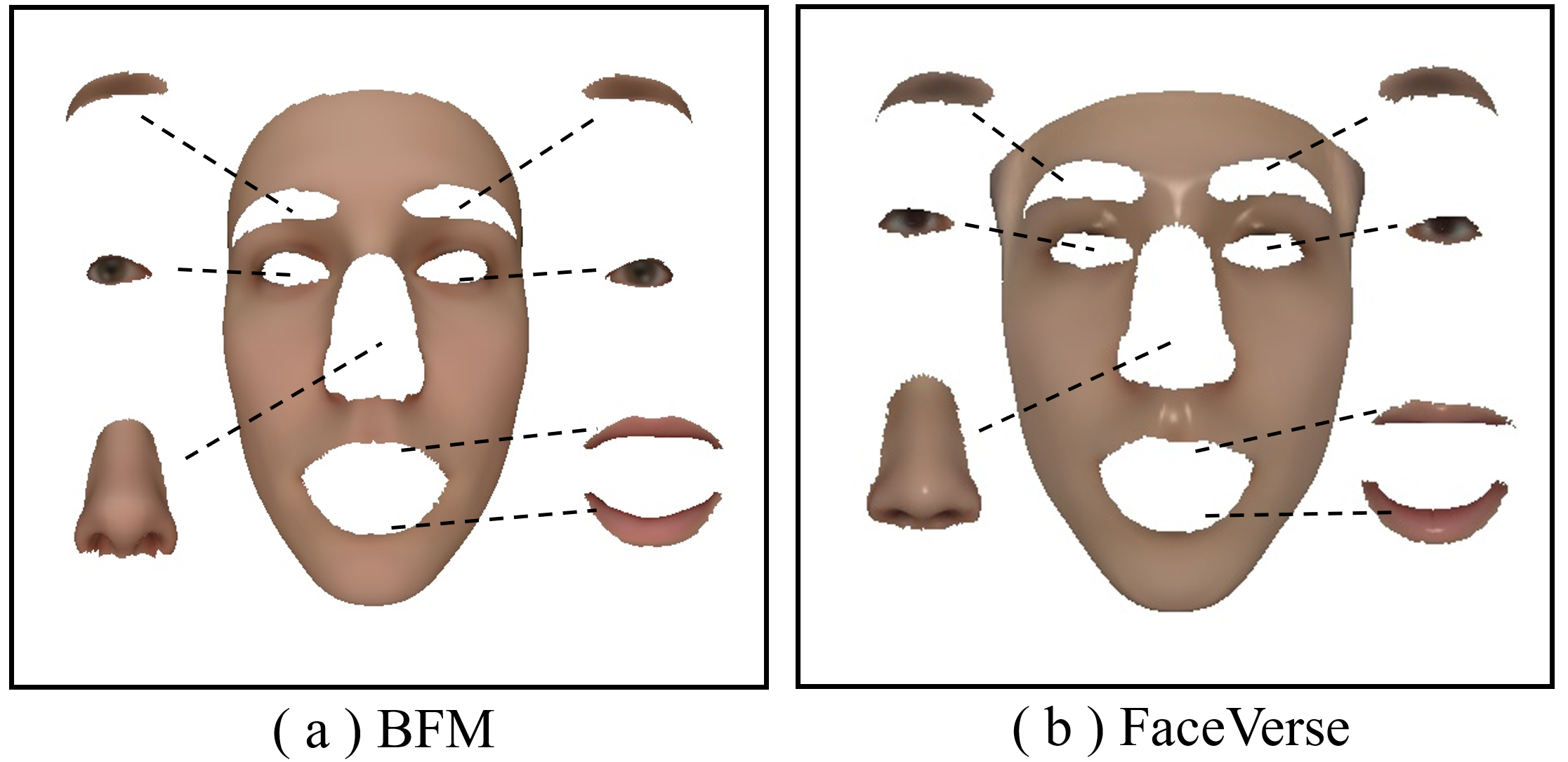 }
\end{center}
\vspace{-0.6cm}
\caption{We provide 3D Mesh part annotations for the BFM \protect\citesupp{paysan20093dsupp} and FaceVerse \protect\citesupp{wang2022faceversesupp} face models, which are well-aligned with the widely recognized 2D face segmentation definitions.}
\vspace{-0.2cm}
\label{sup-anno}
\end{figure}

\myparagraph{3D Mesh Part Annotation.} As shown in the Fig.~\ref{sup-anno}, the objective of $\{Ind_p\}$ is to partition the specific face model to obtain $\{ {V_{2d}^p(\bm{\alpha} )\}} $ that are consistent with the region semantics of 2D segmentation. When ${i \in Ind_p}$, it means that the $i$-th vertex in the mesh belongs to part $p$. 

\begin{figure}[t]
\begin{center}
\includegraphics[width=1\linewidth]{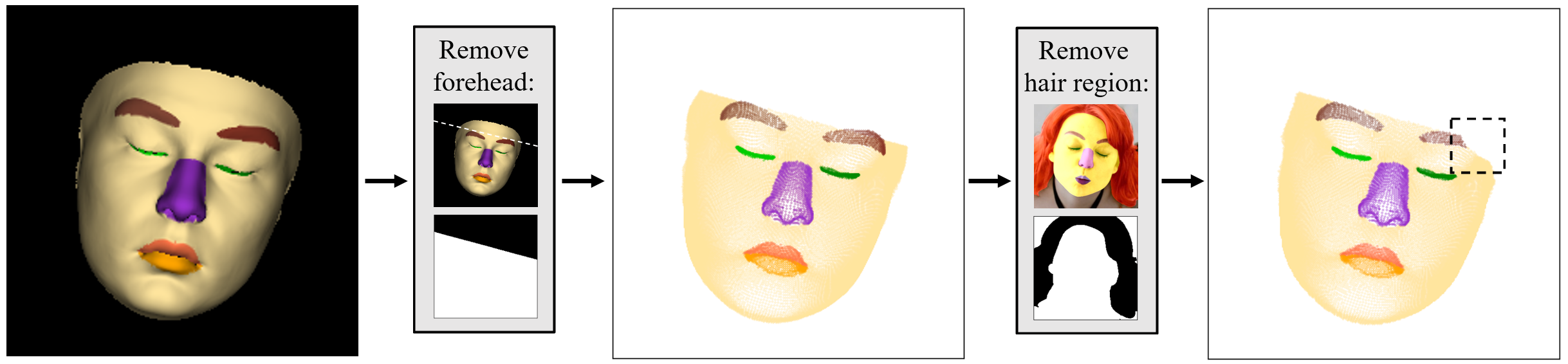 }
\end{center}
\vspace{-0.6cm}
\caption{Remove the forehead region and the points obstructed by hair to ensure consistency with the ground truth $\{{\bm{C}_p}\}$. 
   }
   \vspace{-0.3cm}
\label{remove-v}
\end{figure}

Our 2D to 3D part mesh annotation method is described in Algorithm~\ref{algo} with the following settings: $Render( \cdot )$ renders an image by employing texture on the mesh, and $Seg(\cdot )$ is responsible for segmenting the rendered result. Under the constraint of topological consistency within the same face model, ${V_{3d}}^{all}$ contains 3D face data with distinct poses and expressions, while ${Tex}^{all}$ comprises diverse texture data. ${\bm{P}}=$ \{left\_eye, right\_eye, left\_eyebrow, right\_eyebrow, up\_lip,down\_lip, nose, skin\}. In practice, if the segmentation resolution of the face parsing method is large enough, $k$ could be equal to $1$ in Algorithm~\ref{algo}. The few errant vertex indices in ${\{Ind_p\}}$ should be manually correct. The proposed algorithm~\ref{algo} can also be applied to 2D to 3D landmark marching. To ensure consistency with the ground truth ${{\bm{C}_p}}$, the upper forehead region above the eyebrows is dynamically excluded, and the points obstructed by hair are also removed, as illustrated in Fig.~\ref{remove-v}.

\begin{figure}[t]
\begin{center}
\includegraphics[width=1\linewidth]{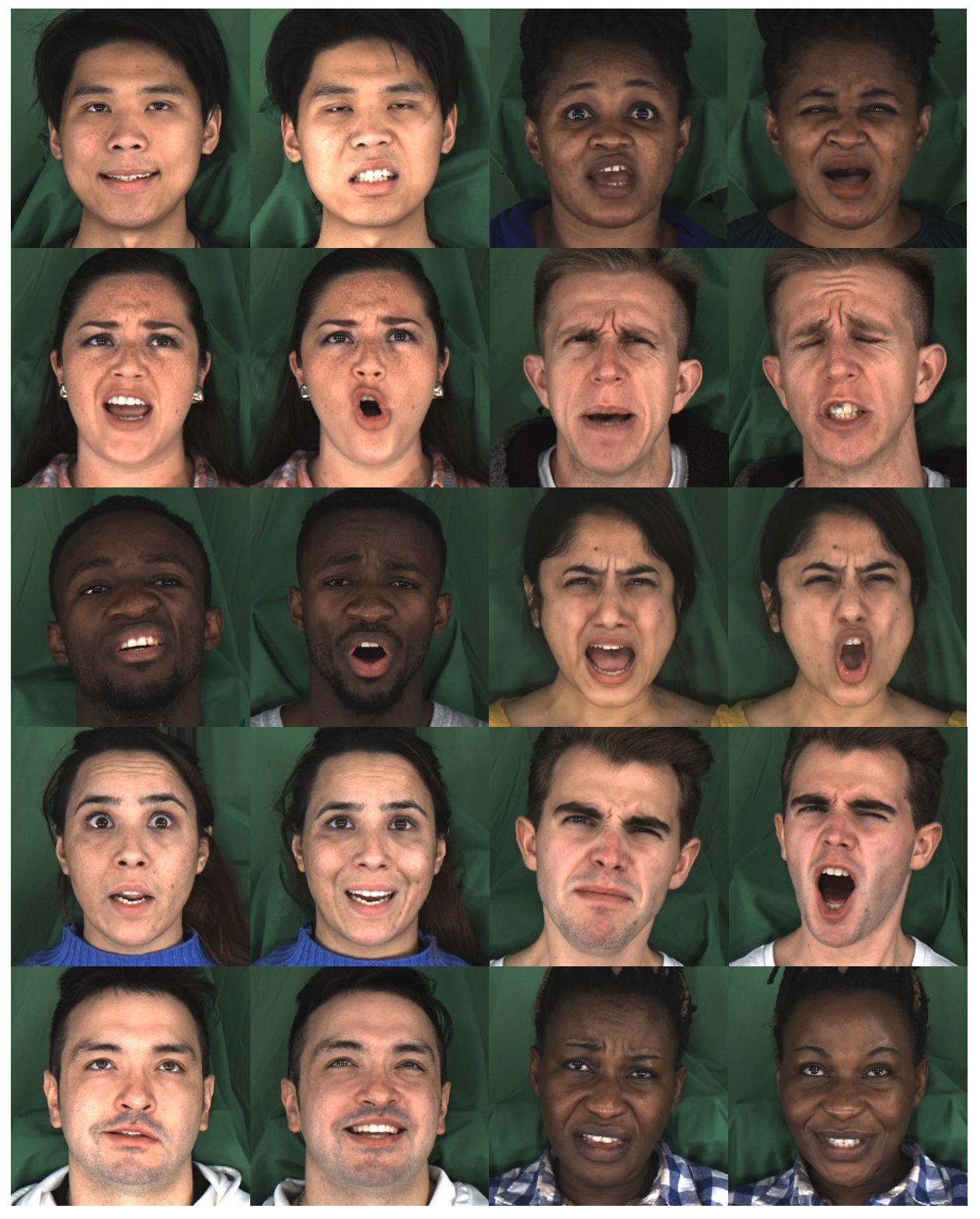 }
\end{center}
\vspace{-0.6cm}
\caption{A subset of test images for Part IoU. 
   }
   \vspace{-0.2cm}
\label{sup-mead}
\end{figure}

\myparagraph{Test Images for Part IoU.} Multi-view Emotional Audio-visual Dataset (MEAD) \citesupp{kaisiyuan2020meadsupp} is a talking-face dataset corpus featuring 60 actors talking with 8 different emotions at three different intensity levels, which can provide high-quality details of facial expressions. We select 10 identities from MEAD, containing diversity across genders and ethnicity. We randomly select 50 different frontal images from each identity to constitute the Part IoU testing set. Fig.~\ref{sup-mead} shows a subset of these images. 

\begin{algorithm}[t]
    \small
        \caption{\small{Identify part indices ${\{Ind_p\}}$ of the mesh.}}
        \label{algo}
        \SetAlgoLined
        \SetKwInput{KwData}{Input}
        \SetKwInput{KwResult}{Init}
         \KwData{$Render( \cdot )$, $Seg(\cdot )$, $V_{3d}^{all}$,${Tex}^{all}$, $\bm{P}$}\vspace{0.00cm}
         \KwResult{${In{d_p} = \varnothing }$,
           $k$ ($k$-nearest-neighbor)
           \vspace{0.00cm}
         }
         \For{$\forall \ {V_{3d}} \in {V_{3d}}^{all}$ and  $\forall \ Tex \in Te{x^{all}}$}{
         \vspace{0.00cm}
            \tcp{Get the segmentation,}
            \vspace{0.00cm}
            ${I^{seg}} = Seg(Render({V_{3d}} , Tex))$,
             \\   \vspace{0.00cm}
         \tcp{Transform ${I^{seg}}$ to coordinates,}
         \vspace{0.00cm}
         $ \{ {\bm{C}_p}|p \in \bm{P}\}  \leftarrow {I^{seg}}$,
         \vspace{0.00cm}
         \\  \vspace{0.00cm}
        \tcp{Project ${V_{3d}}$ to the image plane,}
        \vspace{0.00cm}
         ${V_{2d} = Project( {V_{3d}} )}$
         \vspace{0.00cm}\\
         \For{$p \in \bm{P}$}{
         \vspace{0.00cm}
         \For{$\bm{c} \in \bm{C}_p$}{
         \vspace{0.00cm}
            \tcp{$\bm{c}$ is a 2D coordinate,}
\vspace{0.00cm}
            Find the first $k$ vertices in ${V_{2d}}$ that are closest to $\bm{c}$, and these $k$ vertices should be visible, append their corresponding indices to $Ind_p$.
            \vspace{0.00cm}
     
         }
         }
         }
         \vspace{0.00cm}
         \SetKwInput{KwData}{Output} \KwData{${\{Ind_p\}}$}
\end{algorithm}

\section{More Comparison with the Other Methods}
\begin{figure}[t]
\begin{center}
   \includegraphics[width=1\linewidth]{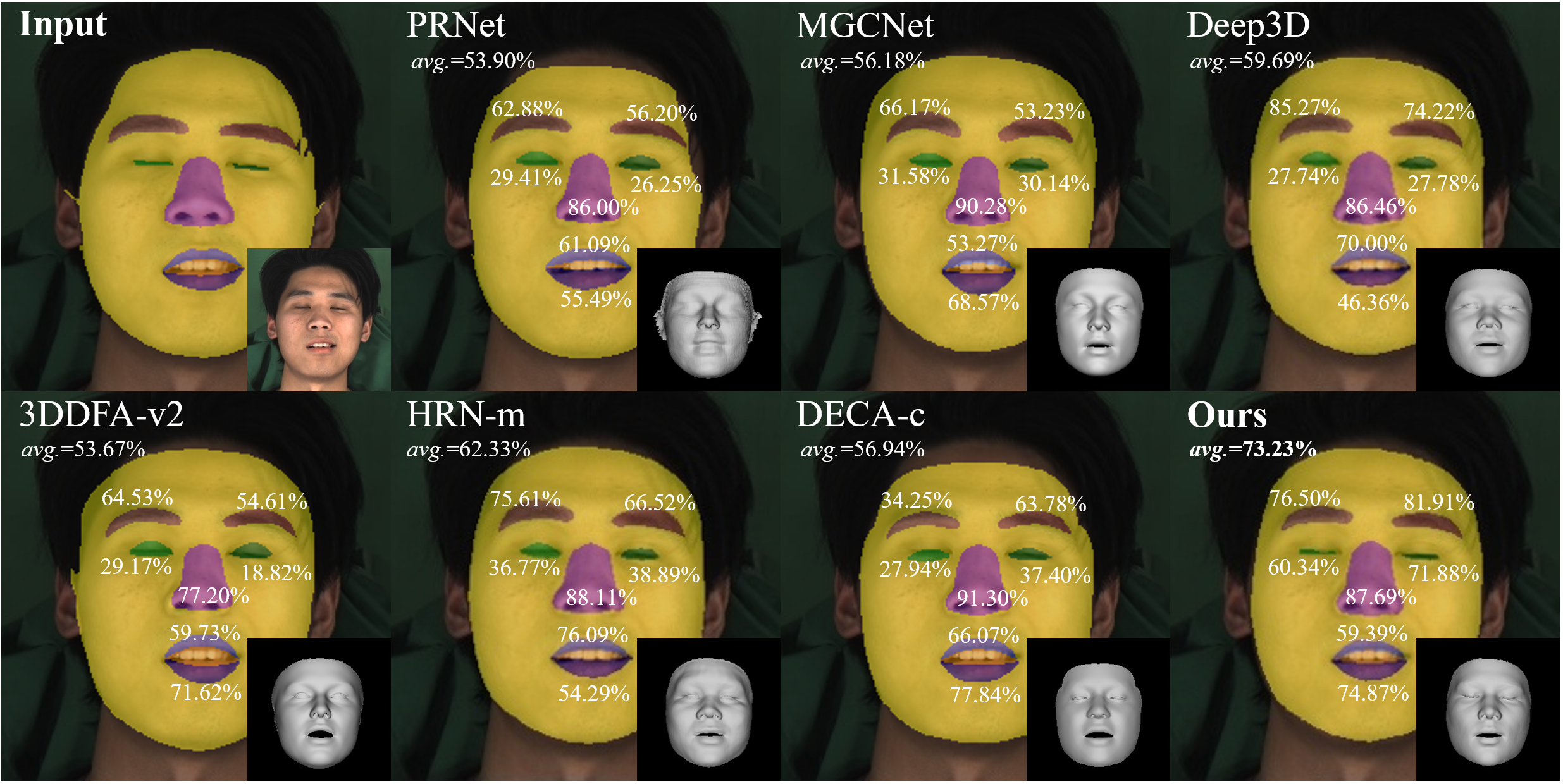}
\end{center}
 \vspace{-0.6cm}
   \caption{Comparison on Part IoU. The IoU value and visualizations for each reconstructed part are annotated, and the bottom right corner of each image is the corresponding 3D reconstruction.
   }
\label{part-iou}
 \vspace{-0.2cm}
\end{figure}

Fig.~\ref{sup-compare} depicts a more comparison between our results and the other state-of-the-art methods, {\ie} PRNet \citesupp{feng2018jointsupp}, MGCNet \citesupp{shang2020selfsupp}, Deep3D \citesupp{deng2019accuratesupp}, 3DDFA-V2 \citesupp{guo2020towardssupp}, HRN \citesupp{lei2023hierarchicalsupp}, and DECA \citesupp{DECA:Siggraph2021supp}. Leveraging the advancements brought by PRDL, our method excel in capturing extreme facial expressions. Part IoU measures the overlap performance between each part of the reconstruction and the ground truth. The visualization of Part IoU for every method can be found in Fig.~\ref{part-iou}, which shows that PRDL enhances the alignment of reconstructed facial features with the original image.

\section{More Results about Synthetic Data}
\begin{figure*}[t]
\begin{center}
\includegraphics[width=1\linewidth]{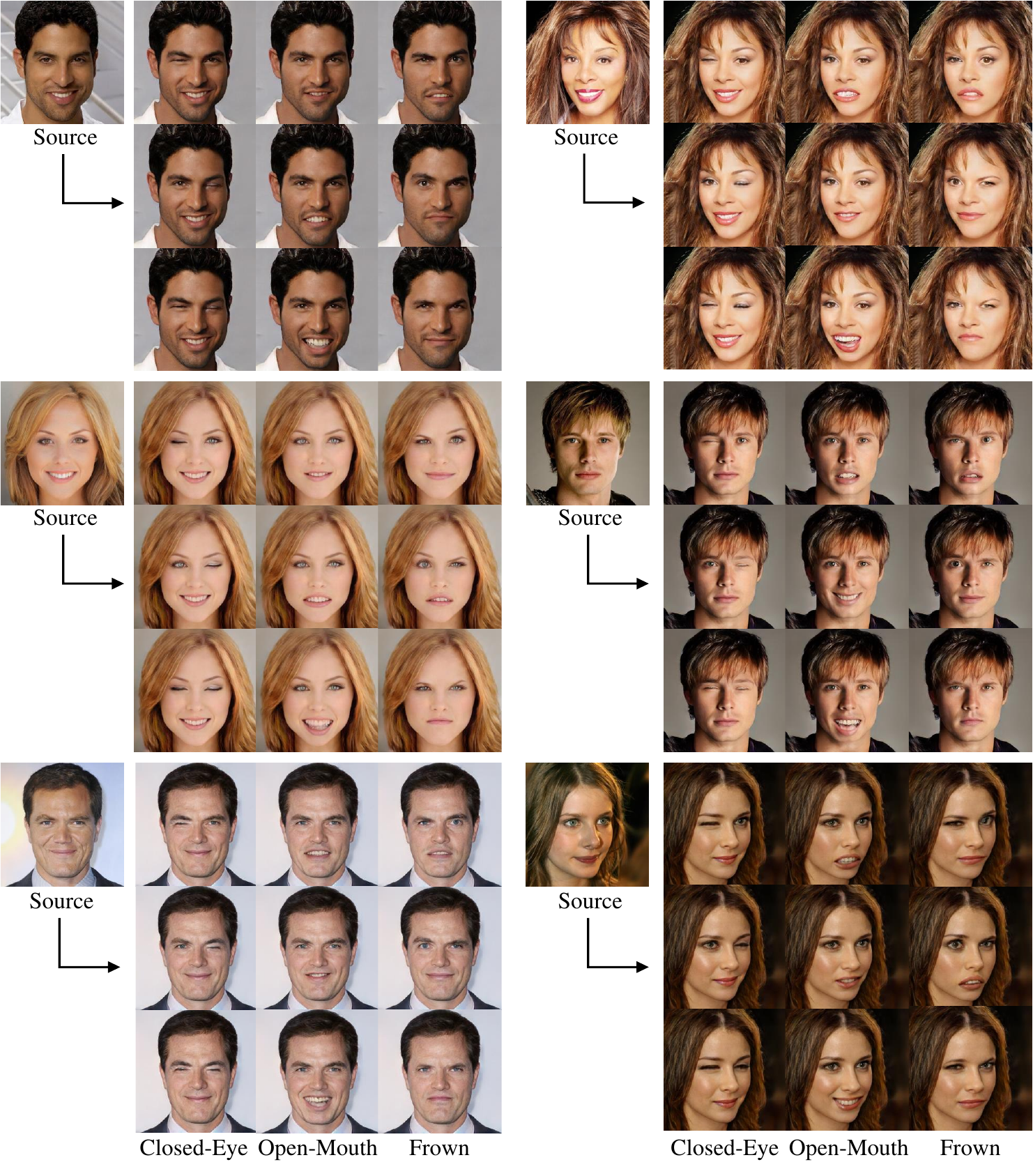}
\end{center}
\vspace{-0.6cm}
\caption{Examples of our synthetic face dataset.
   }
\label{sup-new-data}
\end{figure*}

Fig.~\ref{sup-new-data} illustrates more results about our synthetic emotional expression dataset. The dataset currently consists of over $200K$ images, including synthetic expressions such as closed-eye, open-mouth, and frown. This dataset will be publicly available to facilitate the related research.

\section{Limitations}
\begin{figure}[t]
\begin{center}
\includegraphics[width=1\linewidth]{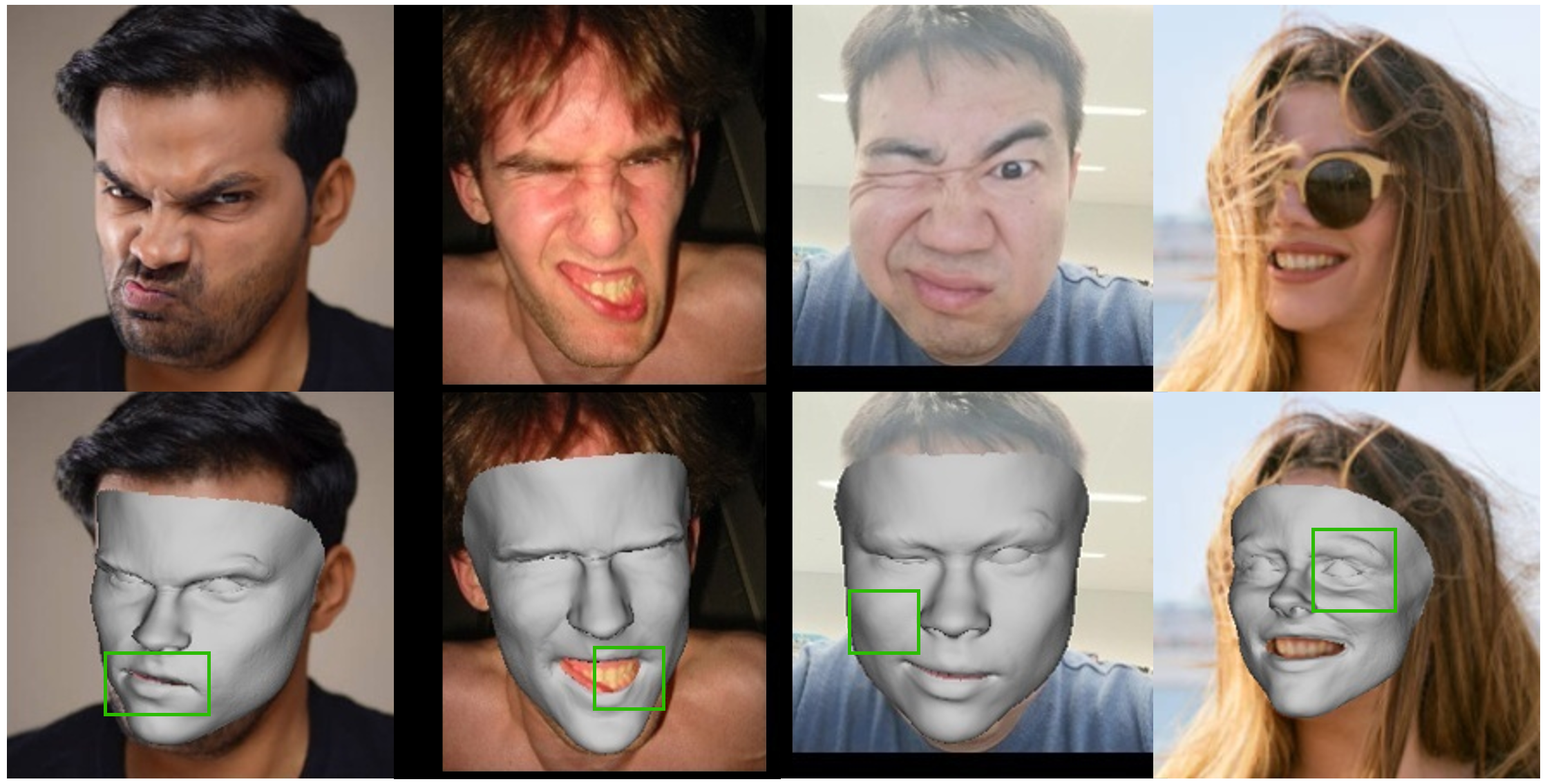 }
\end{center}
\vspace{-0.6cm}
\caption{Limitations of our method. In cases of extremely challenging facial expressions or heavily occluded faces, our reconstructions may exhibit some minor errors.
   }
   \vspace{-0.4cm}
\label{sup-limitation}
\end{figure}

\begin{figure*}[t]
\begin{center}
\includegraphics[width=1\linewidth]{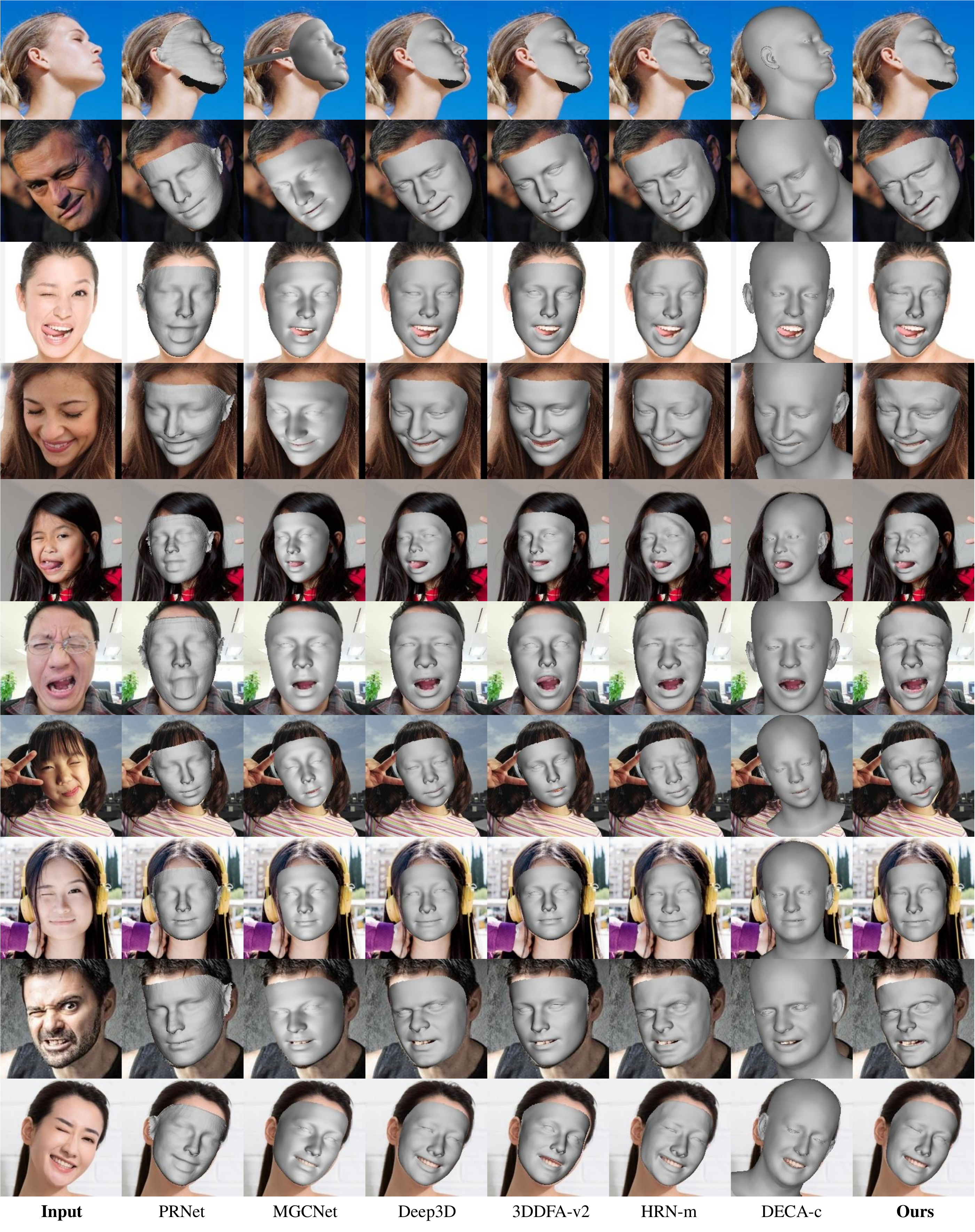 }
\end{center}
\vspace{-0.6cm}
\caption{More comparison with the other methods. From left to right: Input image, PRNet \protect\citesupp{feng2018jointsupp}, MGCNet \protect\citesupp{shang2020selfsupp}, Deep3D \protect\citesupp{deng2019accuratesupp}, 3DDFA-V2 \protect\citesupp{guo2020towardssupp}, HRN \protect\citesupp{lei2023hierarchicalsupp}, DECA \protect\citesupp{DECA:Siggraph2021supp}, and Ours.}
\label{sup-compare}
\end{figure*}

We summarize two limitations of our approach. Firstly, while Fig.~\ref{sup-compare} has demonstrated the excellent performance of our method on extreme facial expressions, it is constrained by the limited linear space of the 3DMM, resulting in some imperfections in reconstructing particularly challenging expressions. Secondly, although our method can handle occluded faces, it may struggle with severe occlusions, as illustrated in Fig.~\ref{sup-limitation}. In the future, we will extend our method to fine-grained face reconstruction and multi-view face reconstruction to address these limitations.

{\small
\bibliographystylesupp{ieeenat_fullname}
\bibliographysupp{prdl_ref_sup}
}

\end{document}